\documentclass[lettersize,journal]{IEEEtran}
\usepackage{amsmath,amsfonts}
\usepackage{algorithmic}
\usepackage{array}
\usepackage[caption=false,font=normalsize,labelfont=sf,textfont=sf]{subfig}
\usepackage{textcomp}
\usepackage{stfloats}
\usepackage{url}
\usepackage{verbatim}
\usepackage{graphicx}
\usepackage{amssymb}
\usepackage{anyfontsize}
\usepackage[T1]{fontenc}
\usepackage{lipsum}
\usepackage{times}
\usepackage{epsfig}
\usepackage{amsmath}
\usepackage{dsfont}
\usepackage{bm}
\usepackage{wrapfig}
\DeclareMathOperator*{\argmax}{arg\,max}

\usepackage{bbding}
\usepackage{booktabs}
\usepackage{multirow}
\usepackage{ragged2e}
\usepackage{tabularx}
\usepackage{subcaption}
\usepackage{color}
\usepackage{amsfonts}

\usepackage[dvipsnames]{xcolor}

\DeclareMathOperator*{\Sharpen}{Sharpen}
 
\usepackage[switch]{lineno}
\usepackage{enumitem}

\usepackage{xcolor}
\usepackage[linesnumbered,ruled,vlined]{algorithm2e}

\SetCommentSty{mycommfont}
\SetKwInput{KwInput}{Input}                % Set the Input
\SetKwInput{KwOutput}{Output}              % set the Output

\def\BibTeX{{\rm B\kern-.05em{\sc i\kern-.025em b}\kern-.08em
    T\kern-.1667em\lower.7ex\hbox{E}\kern-.125emX}}
\usepackage{balance}
\begin{document}
\title{Adaptive Betweenness Clustering for Semi-Supervised Domain Adaptation}

\author{Jichang Li, Guanbin~Li,~\IEEEmembership{Member,~IEEE,}
        and~Yizhou~Yu,~\IEEEmembership{Fellow,~IEEE}% <-this % stops a space
\IEEEcompsocitemizethanks{
\IEEEcompsocthanksitem This work was supported in part by the Guangdong Basic and
Applied Basic Research Foundation (No.2020B1515020048), in part by the
National Natural Science Foundation of China (No.62322608, No.61976250), in part by the Shenzhen Science and Technology Program (NO.~JCYJ20220530141211024). (Corresponding authors are Guanbin Li and Yizhou Yu).
\IEEEcompsocthanksitem J. Li and Y. Yu are with the Department of Computer Science, The University of Hong Kong, Hong Kong (e-mail: csjcli@connect.hku.hk; yizhouy@acm.org).\protect
\IEEEcompsocthanksitem G. Li is with the School of Computer Science and Engineering, Research Institute of Sun Yat-sen University in Shenzhen, Sun Yat-sen
University, Guangzhou, 510006, China (e-mail: liguanbin@mail.sysu.edu.cn).}% <-this % stops a space
%\thanks{Manuscript received July 1, 2021.}
}

\markboth{IEEE TRANSACTIONS ON IMAGE PROCESSING,~VOL. XX, 202X}% 
{Shell \MakeLowercase{\textit{et al.}}: Bare Advanced Demo of IEEEtran.cls for IEEE Computer Society Journals}

\maketitle

\begin{abstract}
Compared to unsupervised domain adaptation, semi-supervised domain adaptation (SSDA) aims to significantly improve the classification performance and generalization capability of the model by leveraging the presence of a small amount of labeled data from the target domain. Several SSDA approaches have been developed to enable semantic-aligned feature confusion between labeled (or pseudo labeled) samples across domains; nevertheless, owing to the scarcity of semantic label information of the target domain, they were arduous to fully realize their potential. In this study, we propose a novel SSDA approach named Graph-based Adaptive Betweenness Clustering (G-ABC) for achieving categorical domain alignment, which enables cross-domain semantic alignment by mandating semantic transfer from labeled data of both the source and target domains to unlabeled target samples. In particular, a heterogeneous graph is initially constructed to reflect the pairwise relationships between labeled samples from both domains and unlabeled ones of the target domain. Then, to degrade the noisy connectivity in the graph, connectivity refinement is conducted by introducing two strategies, namely Confidence Uncertainty based Node Removal and Prediction Dissimilarity based Edge Pruning. Once the graph has been refined, Adaptive Betweenness Clustering is introduced to facilitate semantic transfer by using across-domain betweenness clustering and within-domain betweenness clustering, thereby propagating semantic label information from labeled samples across domains to unlabeled target data. Extensive experiments on three standard benchmark datasets, namely DomainNet, Office-Home, and Office-31, indicated that our method outperforms previous state-of-the-art SSDA approaches, demonstrating the superiority of the proposed G-ABC algorithm.
\end{abstract}


\IEEEPARstart{D}{eep} neural network (DNN) has led to a series of breakthroughs in computer vision tasks such as Image Classification~\cite{mikolajczyk2018data, boulila2021rs, li2019relation, wu2019enhancing, li2022neighborhood}, Semantic Segmentation~\cite{liu2021harmonic, garcia2018, yang2020adversarial, xiong2023unpaired}, Object Detection~\cite{zhao2019object, inoue2018cross, zhao2020collaborative, yan2022unsupervised}, Medical Analysis~\cite{zhou2022generalized, zhou2023transformer}, etc. However, the impressive effectiveness of the training of deep network models remarkably depends on a large number of sample labels, necessitating laborious work in data annotation. An alternative solution comprises boosting the model for the domain of interest (a.k.a., target domain) by employing off-the-shelf labeled training samples from a relevant domain (a.k.a., source domain).
Nonetheless, due to the distribution/domain gap, such a solution often cannot generalize well from the source domain to the target domain to deal with variant circumstances of domain gaps. Unsupervised domain adaptation (UDA), which aims to tackle the distribution gap and decrease the influence of domain shift, has thus gained significant attention for a long time \cite{ealierDA000, ealierDA001, ealierDA002, ealierDA004},~\cite{ganin2016domain, kumagai2019unsupervised, wilson2020survey}. Recently, semi-supervised domain adaptation (SSDA), a variant of the UDA task,  has received wider attention~\cite{saito2019semi, kim2020attract, li2021cross}. With a few labeled target samples, SSDA can significantly enhance the adaptation model's performance w.r.t the target domain, compared to unsupervised domain adaptation. In this way, a small amount of annotated data in the target domain can be used to expand the semantic space, allowing a large number of samples of the same category from different domains to be clustered together at the feature level, so as to achieve partial categorical alignment.

\begin{figure}[t]
\centering
\includegraphics[width=9.0cm,height=5.0cm]{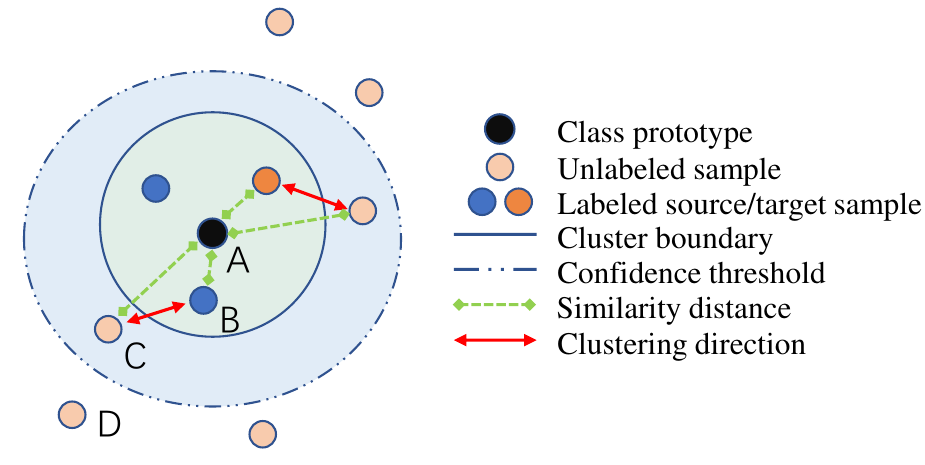} 
\caption{An example to illustrate Adaptive Betweenness Clustering (ABC). The proposed G-ABC algorithm conducts sample clustering between a labeled point (e.g., ``B'') and an unlabeled point (e.g., ``C''), when they have similarity distances within a confidence threshold to the same class prototype (e.g., ``A'') and they are with similar prediction distributions from the model. Herein, Point ``A'' guides the clustering process by serving as an intermediary (betweenness) point for Points ``B'' and ``C''. Point ``D'' is outside of the clustering range. 
}
\label{Figure-Illustration}
\end{figure}

Despite the advantages of SSDA over UDA, SSDA also presents its own specific challenges.
At first, in the SSDA scenario, a supervised model trained on a small number of labeled target samples and a large amount of labeled source data can only achieve partial cross-domain feature alignment, as it only aligns features of labeled target samples and their correlated nearby samples with the corresponding feature clusters in the source domain~\cite{kim2020attract}. In addition, the trained SSDA model is incapable of producing highly discriminative feature representations for the target domain since the massive labels of the source domain dominate the supervision and causes the learned feature representations to be biased towards the discrimination of the source domain~\cite{saito2019semi}. Preliminary SSDA works, such as~\cite{saito2019semi, kim2020attract, li2021cross,  qin2021contradictory}, have each proposed their individual solutions to tackle these challenges, and significant performance improvement has been witnessed. However, as with previous UDA studies focusing on global feature confusion at the domain level~\cite{pan2010domain, yan2017mind, ganin2016domain, tzeng2017adversarial}, existing works are still unable to reach globally categorical domain alignment due to the scarcity of semantic label information for each category in the target domain. In other words, despite the presence of perfect domain-level alignment in feature confusion, it leads to label mismatch between massive unlabeled target data and the data of the source domain, hence compromising the model's performance.  Recent SSDA algorithms, such as~\cite{li2021ecacl, luo2021relaxed}, have demonstrated that the semantic-aligned feature confusion across domains appears to work properly in semi-supervised domain adaptation, since during domain alignment, sample features from both domains with the same class will likely be aggregated into a same cluster. However, these methods achieve categorical domain alignment mostly through promoting semantic alignment between labeled samples across domains, with far less emphasis on employing a vast amount of unlabeled target samples.

In this paper, we present a novel SSDA approach, named {\bf G}raph-based {\bf A}daptive {\bf B}etweenness {\bf C}lustering ({\bf G-ABC}), to tackle the challenges of the SSDA tasks. To achieve globally categorical domain alignment, we propose to enforce semantic transfer from labeled samples across domains to unlabeled target samples in order to promote cross-domain semantic alignment. Using the ground-truth sample labels from both domains as references, the trained SSDA model may thereby propagate semantic label information to the unlabeled target samples. 
In this way, a substantial amount of target label information is augmented through semantic propagation, thus significantly enhancing the generalization of the model to the target domain.
Specifically, we first construct a heterogeneous graph to capture the pairwise associations between unlabeled target samples and labeled samples from either the source or target domain, based on the pairwise label similarity of those paired samples. Then, we provide two connectivity refinement strategies, namely Confidence Uncertainty based Node Removal and Prediction Dissimilarity based Edge Pruning, to eliminate the noisy connectivity in the graph. In detail, the former degrades the connectivity towards unreliable unlabeled samples by removing unlabeled target instances with low predicted confidence, whereas the latter prunes the graph connections between samples with divergent probabilistic prediction distributions.
 
With the refined graph structure, we design a new clustering algorithm, namley Adaptive Bewteenness Clustering (ABC) as shown in Fig.~\textcolor{black}{\ref{Figure-Illustration}}.  To achieve semantic transfer, in this algorithm, we model the task of sample clustering between a labeled and an unlabeled sample as a binary pairwise-classification problem. The fundamental premise behind such an algorithm is to aggregate the feature representations of the paired samples that share the same class in the graph while separating those of different classes. In particular, this algorithm involves two strategies to achieve semantic propagation, namely across-domain betweenness clustering (ADBC) and within-domain betweenness clustering (WDBC), by clustering the unlabeled target samples towards the labeled source or target domains. As a result, the ADBC strategy encourages alignment between unlabeled target samples and the source domain, whereas the WDBC scheme strengthens linkages between labeled and unlabeled target samples. Ultimately, semantic label information can be gradually transferred into unlabeled target instances as model training iterates. In this way, the rising balance of semantic label information of the source and target domains eliminates model bias toward the source domain and achieves globally categorical domain alignment, driving the model to generate more domain-invariant yet discriminative target features.

To sum up, our main contributions can be shown as follows:
\begin{itemize}

    \item We propose a novel SSDA framework called {\bf G}raph-based {\bf A}daptive {\bf B}etweenness {\bf C}lustering ({\bf G-ABC}) to tackle semi-supervised domain adaptation. To achieve globally categorical domain alignment, the proposed G-ABC conducts cross-domain semantic alignment with semantic transfer from labeled samples of both domains to unlabeled target data.
    
    \item We construct a heterogeneous graph to characterize the associations between unlabeled target examples and labeled data of both domains. Two connectivity refinement strategies, namely Confidence Uncertainty based Node Removal and Prediction Dissimilarity based Edge Pruning, are further provided to decrease the noisy connectivity in the graph.
    
    \item Given the above-refined graph structure, we propose Adaptive Betweenness Clustering to impose semantic transfer across domains;  in particular, we design across-domain betweenness clustering and within-domain betweenness clustering, respectively, to propagate semantic label information from labeled source and target domains to unlabeled target samples.
    
    \item We perform extensive experiments on three standard benchmark datasets, including DomainNet~\cite{peng2019moment}, Office-Home~\cite{venkateswara2017deep} and Office-31~\cite{saenko2010adapting}, to verify the effectiveness of our proposed method, and the results show that our method outperforms all previous state-of-the-art SSDA methods by clear margins.

\end{itemize}

This paper is organized as follows. In Sec.~\ref{Section:RelatedWork}, we overview the prior research related to our work. In Sec.~\ref{Section:Methodology}, we present and describe the proposed algorithm for semi-supervised domain adaptation. Furthermore, we conduct comparative experiments to evaluate the performance of the proposed approach in Sec.~\ref{Section:Experiment}, while the conclusions are illustrated in Sec.~\ref{Section:Conclusions}.
 
\section{Related Work}
\label{Section:RelatedWork}

\subsection{Domain Adaptation}

Domain adaptation (DA) addresses the problem of generalizing a model trained on a large number of labeled samples from the source domain to the target domain~\cite{huang2023divide, xu2019larger, zhuang2022discovering, zhang2022divide, zhang2022divide}. With the goal of decreasing the distribution gap across various domains, the most challenging issue of the DA problem is assisting the model in learning domain-invariant features. To accomplish domain adaptation, early classic algorithms to handle the domain adaptation tasks involve reducing the distribution discrepancies across domains assessed by Maximum Mean Discrepancy (MMD)~\cite{pan2010domain, yan2017mind}, sharing the identical cross-domain statistics (e.g., mean value and covariance)~\cite{sun2016deep}, and so on. Tzeng et al. in~\cite{tzeng2014deep}, for instance, presented a method for leveraging MMD to drive the model to generate domain-invariant features by assessing the discrepancies between the model outputs of both domains. Long et al. optimized the adaptation model from a different perspective~\cite{long2015learning}, i.e., employing multi-kernel MMD to evaluate the output differences of samples across domains at multiple model levels.
In addition, recent advances trends favor adversarial domain alignment to expedite feature alignment across domains so that knowledge from classifiers trained on labeled source samples can be efficiently transferred to the target domain~\cite{ganin2016domain, tzeng2017adversarial, su2020active, chen2020adversarial, cao2018partial, awais2021adversarial, yang2021exploring}.
For example, Saito et al. conducted minimax training over unlabeled target samples to cluster these target features around domain-invariant class prototypes, thus imposing cross-domain feature alignment~\cite{saito2019semi}. The aforementioned DA algorithms aimed at aligning source and target features at the domain level. However, many related advances, such as \cite{chen2019progressive, pan2019transferrable, li2021ecacl, R2C1}, demonstrated that decreasing the discrepancies of conditional distributions towards categorical domain alignments is preferred, resulting in improved adaptation between domains. 
To this end, it should be natural to incorporate semantic label information into adaptation. For instance, semantic alignment was proposed in~\cite{motiian2017few, li2021semantic, xiao2021implicit} to achieve this. Motivated by it, we also emphasize semantic-level domain alignment, leveraging source and target labels as references, and encouraging semantic transfer from labeled source and target domains to unlabeled target samples.
 
\begin{figure*}[t]
\centering
\includegraphics[width=18.0cm,height=5.5cm]{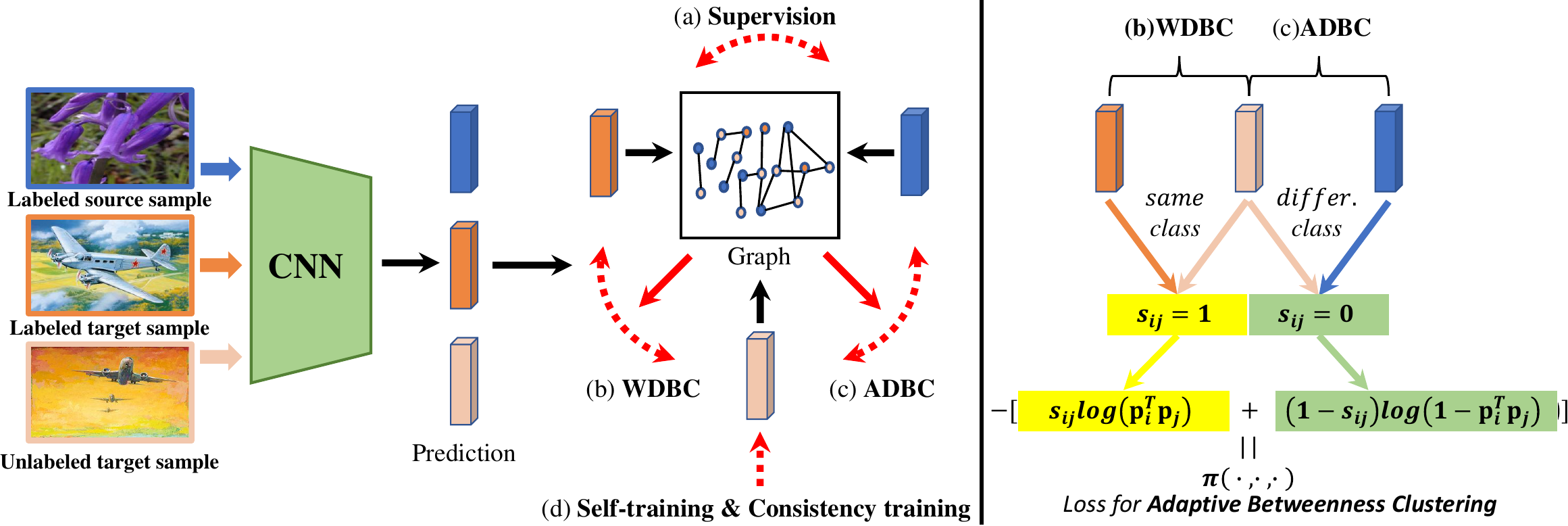} 
\caption{An overview of the proposed framework and the training loss for Adaptive Betweenness Clustering. \textbf{Left: }(a) Supervision of labeled data from both source and target domains is applied to guarantee partially categorical domain alignment. (b) Within-domain betweenness clustering (WDBC) is used to determine the relationship between labeled and unlabeled target data. (c) Across-domain betweenness clustering (ADBC) is used to effectively align unlabeled target samples with the source domain. D) Auxiliary techniques for model optimization, including self-training, consistency training, etc. These four components together enable globally categorical domain alignment, progressively enhancing the model's performance, with (b) and (c) establishing reliable sample connectivity among training samples, represented by a heterogeneous graph. \textbf{Right:} Given a pairwise label $s_{ij}$ between samples, the training loss of Adaptive Betweenness Clustering aims to bring samples from the same class closer together in the feature space when $s_{ij}=1$, or to separate samples from different classes when $s_{ij}=0$. This allows for semantic transfer from labeled source or target domains to unlabeled target samples.
Note that the orange and light-orange samples belong to the same category, namely ``plane'', while the blue sample is from a different category, i.e. ``flower''.}
\label{Figure:Framework}
\end{figure*}

\subsection{Domain Adaptation Related to Graphs}

The graphs employed in domain adaptation capture latent topological structures among the training data across domains, such that the learned relationships between domain samples can then be leveraged to encourage the model to better adapt reliable data structures from both domains.  In general, samples from the source and target domains are represented by distinct graph structures.
Thus, many previous DA algorithms relating to graphs, such as~\cite{zhang2019dane, wu2020unsupervised, yang2021domain}, first attempted to transfer knowledge learned on a labeled source graph to an unlabeled target graph.  Based on the labeled graph in the source domain, these algorithms engaged source labels as the supervision signals, and the model therefore received training on both the source graph and the target graph respectively. For example, Pilanci {et al.} employed frequency analysis to align two data graphs through which information can be transferred or shared~\cite{pilanci2019domain, pilanci2020domain}. On the other hand, Ding {et al.} in~\cite{ding2018graph} proposed constructing a cross-domain graph based on samples from both domains to capture the intrinsic structure in the shared space among the training data in order to concurrently enforce domain transfer and label propagation.
In this way, domain-invariant feature learning and target discriminative feature learning are unified into the same framework, thus benefiting each other for more effective knowledge transfer. Based on such an observation, we also propose to hire the cross-domain graph to achieve categorical domain alignment. Nevertheless, earlier related works usually model graph Laplacian regularization to push graph nodes closer, but this unsupervised technique disregards the usage of sample labels to assist semantics alignment~\cite{zhou2003learning, nguyen2011discriminative, wang2018adaptive}. In our work, we design a novel clustering algorithm called Adaptive Betweenness Clustering to take full advantage of both source and target labels, thereby contributing to enhanced performance for the model to classify target samples.
 
\subsection{Semi-supervised Domain Adaptation}

Due to the availability of a few target labels, semi-supervised domain adaptation has the potential to significantly improve the classification performance of the model on the target domain in comparison to unsupervised domain adaptation~\cite{kim2020attract, li2020online, chen2020semi, YangLuyu2020DCwT, R2C2}. Recent progress in SSDA, such as~\cite{saito2019semi, kim2020attract, 2020Bidirectional, li2020learning, qin2021contradictory, li2021cross} have primarily focused on adversarial training to align cross-domain feature distributions. Here we mainly describe some related SSDA approaches that do not involve adversarial learning. For example, Samarth et al. in~\cite{mishra2021surprisingly} demonstrated that without the need for conventional adversarial domain alignment, self-supervision based pre-training and consistency regularization might be relied upon to produce a stronger classifier in the target domain, while Luo et al. in~\cite{luo2021relaxed} developed ``Relaxed cGAN'' to transfer image styles from source samples to unlabeled target samples in order to help achieve domain-level distribution alignment. Besides, Yoon et al. in \cite{R2C2} also focus on style transferring for achieving better adaptation, generating assistant features by transferring intermediate styles between labeled and unlabeled samples.

In addition to bridging the gap and exchanging knowledge between the source and target domains, Yang et al. in~\cite{YangLuyu2020DCwT} proposed decomposing the SSDA task into a semi-supervised learning (SSL) problem and an unsupervised domain adaptation problem. Specifically, the former is used to improve discrimination in the target domain, whereas the latter facilitates domain alignment. Such an algorithm trains two distinct classifiers utilizing Mixup and Co-training, respectively, in order to learn complementary features to each other, resulting in better domain adaptation. Similarly,~\cite{R2C1} also performed adaptation by learning two classifier networks, trained for contradictory purposes. The first classifier groups target features to enhance intra-class density and increase categorical cluster gaps for robust learning, while the second, as a regularizer, disperses source features for a smoother decision boundary.

In this paper, we extend the motivation from~\cite{luo2021relaxed, li2021ecacl} and propose G-ABC, which achieves categorical domain alignment by providing increased access to unlabeled target samples during adaptation. 

While there appears to be a superficial similarity between \cite{R1C1} and our proposed method in encouraging consistent predictions between features, we would like to clarify their differences as follows.

\begin{enumerate}

    \item \cite{R1C1} merely encourages prediction consistency or similarity between a feature and its few neighbors, all of which are from the unlabeled target data. In contrast, our method, G-ABC, employs the proposed clustering technique called Adaptive Betweenness Clustering to group unlabeled samples toward labeled source or target instances. This is achieved by enforcing consistent probabilistic prediction distributions between two similar samples while forcing inconsistency otherwise. More specifically, using the ground-truth sample labels from both domains as references, G-ABC is more effective in propagating the semantic label information from the labeled source and target domains to the unlabeled target examples. Conversely, \cite{R1C1} cannot achieve this, as it only makes the unlabeled target features more compact in an unsupervised manner.

    \item The pairwise label similarity we propose can be viewed as a more credible measure of pairwise relationships than the pairwise feature similarity used in {\cite{R1C1}}. This is especially true when the label information of labeled examples across domains can be trusted. Building upon this premise, we have taken two connectivity refinement strategies to build a reliable graph structure of pairwise relationships, which effectively mitigates the potential harm of noisy connectivity on model performance improvement. In contrast, the pairwise feature similarity introduced in {\cite{R1C1}} involves unsupervised sample matching, which makes it more challenging to accurately pair samples of the same class.

\end{enumerate}
 
\section{The Proposed Method}
\label{Section:Methodology}

In SSDA, we are given labeled samples from the source and target domains, denoted by $\mathcal{D}_{s}={\{(x_i^s,y_i^s )\}}_{i=1}^{N_s }$ and $\mathcal{D}_{l}={\{(x_i^l,y_i^l)\}}_{i=1}^{N_l}$, as well as unlabeled samples from the target domain, denoted as $\mathcal{D}_{u}={\{(x_i^u, )\}}_{i=1}^{N_u}$, where $N_s$, $N_l$ and $N_u$ are the sizes of $\mathcal{D}_{s}$, $\mathcal{D}_{l}$ and $\mathcal{D}_{u}$, respectively. Our goal is to train an SSDA model using $\mathcal{D}_{s}$, $\mathcal{D}_{l}$ and $\mathcal{D}_{u}$, followed by evaluating the trained model on the target domain.

Like existing SSDA works, such as~\cite{saito2019semi, 2020Bidirectional, kim2020attract}, we first parameterize the SSDA model by $\theta$, made up of two components, namely a feature extractor $\mathcal{F}$ and a classifier $\mathcal{G}$. 
The classifier $\mathcal{G}$ is an unbiased linear network with a normalization layer that maps the extracted features from the feature extractor $\mathcal{F}$ into a spherical feature space. Here, the weight vectors of the classifier are denoted as $W=[w_1,w_2,\cdots,w_K]$, and these vectors can be regarded as the prototypes that represent $K$  classes~\cite{saito2019semi, saito2020universal}. Accordingly, samples with the same class label from the source or target domains are mapped nearby to the same class prototype in the feature space. As demonstrated in~\cite{saito2019semi, kim2020attract}, this has a considerable impact on minimizing the cross-domain feature variance of samples with the same class label. In a short, the normalized feature with temperature $T$ of an input image $x$, $\mathbf{f}=\frac{1}{T}\frac{\mathcal{F}(x)}{\|\mathcal{F}(x)\|}$, is fed into the classifier $\mathcal{G}$ to obtain the probabilistic prediction as follows:
\begin{equation}
p(x)=\sigma(\mathcal{G}(\mathbf{f}))=\sigma(W^{\mathsf{T}}\mathbf{f}),
\end{equation}
where $\sigma\left(\cdot\right)$ is a softmax function. $p(x)$ reflects the similarity scores, achieved by calculating the cosine distances, between the point $x$ and the prototypes of distinct classes. For convenience, we often abbreviate $p(x)$ as $\mathbf{p}$, i.e., $\mathbf{p}=p(x)$. 

\noindent\textbf{Overview.}
In this paper, we propose Graph-based Adaptive Betweenness Clustering (G-ABC) to tackle semi-supervised domain adaptation. In detail, we first construct a heterogeneous graph to depict the pairwise associations between labeled samples from both domains and unlabeled examples from the target domain. Then, to degrade the noisy sample connectivity, we refine the original heterogeneous graph using two strategies: Confidence Uncertainty based Node Removal (CUNR) and Prediction Dissimilarity based Edge Pruning (PDEP). Afterwards, to associate complementary characteristics of the source and target labels with unlabeled target samples, we present Adaptive Betweenness Clustering to conduct semantic propagation, facilitating semantic alignment between domains. Finally, we leverage off-the-shelf and well-established techniques, such as pseudo-label selection, self-training~\cite{arazo2020pseudo, sohn2020fixmatch} and consistency training~\cite{xie2019unsupervised, li2019semi, wu2019mutual}, to further optimize the model in order to achieve globally categorical domain alignment. An overview of the proposed method has been summarized in Fig.~\ref{Figure:Framework}.
 
\begin{figure*}[t]
\centering
    \includegraphics[width=18.0cm,height=7.5cm]{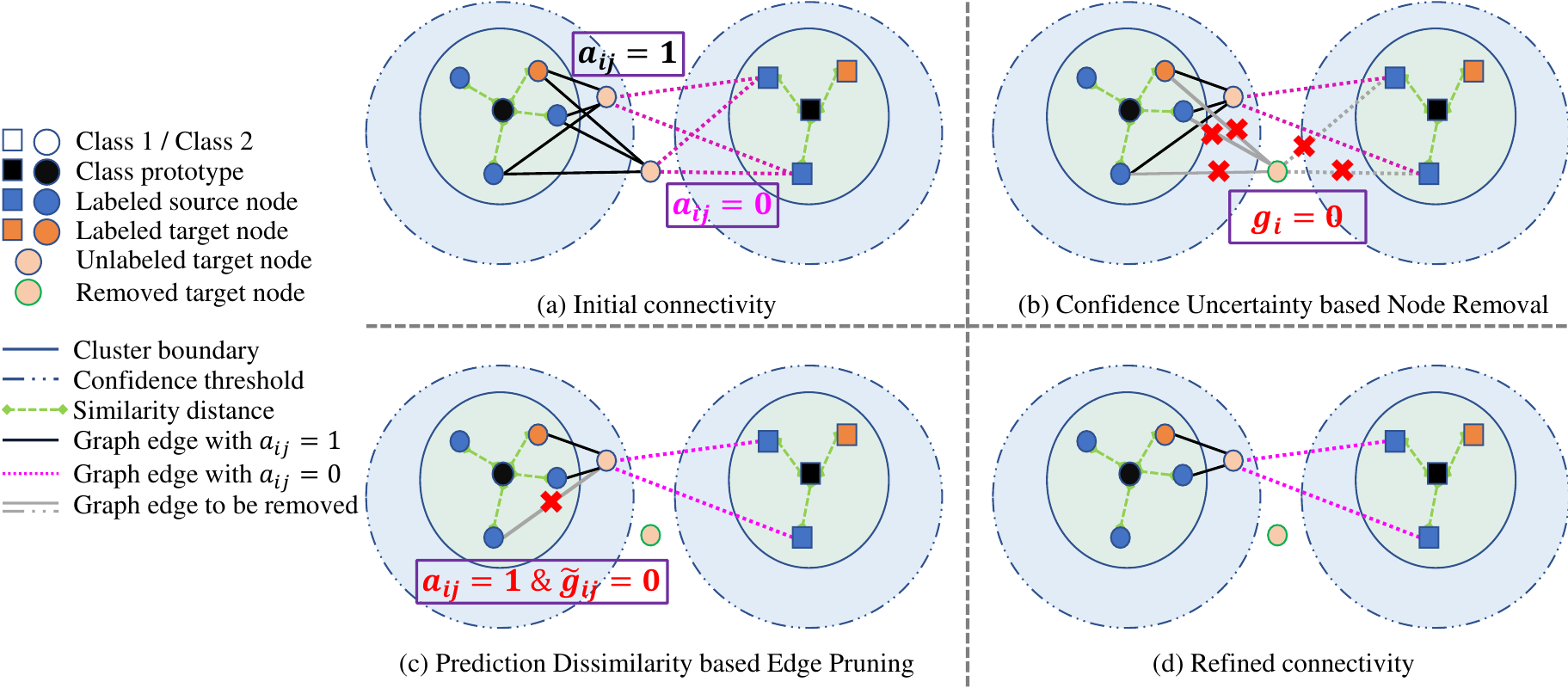} 
    \caption{
    A diagram depicting graph construction and connectivity refinement. (a) demonstrates the initially constructed sample connectivity of the graph, while (d) presents the refined graph after the connectivity refinement process is performed. The connectivity refinement process is further illustrated by (b) and (c), which effectively eliminate noisy connectivity through the CUNR and PDEP strategies, respectively, resulting in a more reliable graph structure to represent the relationships between samples. The technical details of these four subdiagrams are as follows: (a) Using pairwise label similarities between samples, the initial connectivity between training examples in a heterogeneous graph is constructed; (b) Confidence Uncertainty based Node Removal (CUNR) reduces the connectivity towards unreliable unlabeled samples by removing nodes with low predicted confidence; (c) Prediction Dissimilarity based Edge Pruning (PDEP) further removes the connections between graph samples whose probabilistic prediction distributions are dissimilar; (d) A refined graph is obtained to properly capture the pairwise associations between samples.}
    \label{Figure:Diagram}
\end{figure*}

\subsection{Graph Construction}

The goal of graph construction is to discover the sample connectivity of the training data with a heterogeneous graph $G =\langle V, E, A \rangle$. In particular, $V=\{v_i\}_{i=1}^{N_s+N_l+N_u}$ represents the collection of graph nodes consisting of labeled source instances from~$\mathcal{D}_{s}$, labeled target samples from~$\mathcal{D}_{l}$, and unlabeled target samples from~$\mathcal{D}_{u}$, whereas $E$ collects pairwise associations between a graph node of the unlabeled samples and the other node of the labeled examples. Then, the relationships between graph nodes given by the non-negative affinity matrix $A$ can be calculated as follows,
\begin{equation*}
\label{matrix}
A = 
\begin{bmatrix}
a_{1,1}     & \cdots & a_{1,N_{l}}     & a_{1,{N_{l}}+1}     & \cdots & a_{1,{N_{l}}+{N_{s}}}   \\
a_{2,1}     & \cdots & a_{2,N_{l}}     & a_{2,{N_{l}}+1}     & \cdots & a_{2,{N_{l}}+{N_{s}}}   \\
\vdots      & \ddots & \vdots          & \vdots              & \ddots & \vdots                  \\
a_{N_{u},1} & \cdots & a_{N_{u},N_{l}} & a_{N_{u},{N_{l}}+1} & \cdots & a_{N_{u},{N_{l}}+{N_{s}}}
\end{bmatrix},
\end{equation*}
where rows of this matrix denote the unlabeled samples of the target domain, while the first $N_l$ columns and the last $N_s$ columns of $A$ refer to the labeled samples from the target and source domains, respectively.
In addition, $a_{ij}$ (i.e., the $ij$-th entry of $A$) is the weight of the edge connecting between graph nodes $v_i$ and $v_j$, which encodes the mutual relationship of the sample pair. 

Generally, the weight of a connectivity in a graph can be determined using cosine similarity between sample features~\cite{li2018heterogeneous, bai2021hierarchical}.
We propose in this paper that the pairwise associations can be established by comparing the given ground-truth labels of labeled samples to the model's predicted class labels of unlabeled samples; we refer to this as pairwise label similarity. 
Compared to pairwise feature similarity, pairwise label similarity can be viewed as a more credible measure of pairwise relationships, providing that the label information of labeled examples across domains is trustworthy. In addition, for the predicted label of the unlabeled target sample, we take measures to mitigate the noisy connectivity of the established graph (See Sec.~\ref{SubSection:Refinement}).
Afterwards, we can obtain the pairwise label similarity $a_{ij}$ of an edge between graph nodes consisting of an unlabeled sample $x_i$ ({a.k.a}, $v_i$), and a labeled sample $x_j$ ({a.k.a}, $v_j$) affiliated with its ground-truth class label $y_j$, as follows,
\begin{equation}
    \label{graph_w}
    a_{ij} =  \mathds{1}\{{\hat{y}_i=y_j}\}, 
\end{equation}
where $\hat{y}_i={\argmax}_{k}{(p(x_{i})[k])}$ is the predicted class label of the unlabeled target sample $x_i$, while $\mathds{1}\{\cdot\}$ is a binary indicator function. Notice that we illustrate the initial sample connectivity of the built graph in Fig.~\textcolor{black}{\ref{Figure:Diagram}}(a).

\subsection{Connectivity Refinement for Graph Unreliability}
\label{SubSection:Refinement}

Once we obtain the initial heterogeneous graph, we introduce connectivity refinement to alleviate the unreliability of the graph.
Due to high uncertainties caused by low predicted confidence, the graph nodes corresponding to unlabeled target samples are susceptible to receiving erroneous pseudo-labels, thereby resulting in noisy connectivity of the graph. In addition, as demonstrated in~\cite{wang2007label, gong2016label, TPN2019}, labels and features should vary smoothly over the edges of the graph so as to well conduct semantic propagation. To this end, features between nodes $v_i$ and $v_j$ of an edge should have a high degree of similarity when $a_{ij}=1$, i.e., they should be neighbors in the spherical feature space. 
In this situation, the probabilistic prediction distributions of these two node samples predicted by the classifier ought to be similar. Consequently, there would be additional edges with noise in the graph, for which the probabilistic prediction distributions of both node samples from the model have a very low similarity. To eliminate the connectivity with noise and unreliability in the graph $G$, we present two connectivity refinement strategies, namely Confidence Uncertainty based Node Removal and Prediction Dissimilarity based Edge Pruning, resulting in a refined graph that represents the pairwise relationship of the data structure with high reliability. An illustration is shown in Fig.~\textcolor{black}{\ref{Figure:Diagram}}(b), (c), and (d).

\noindent\textbf{Confidence Uncertainty based Node Removal (CUNR).}
This strategy degrades the connectivity towards unreliable unlabeled samples through the removal of unlabeled target instances with low-confidence predictions. We employ a sufficiently high confidence threshold $\tau \in [0, 1]$ to choose reliable candidate nodes from unlabeled target samples:
\begin{equation}
    \label{graph_gi1}
    g_i =  \mathds{1}\{{\max}_{k}{(p(x_{i})[k])}>\tau\}, 
\end{equation}
where $g_i$ is a binary indicator to preserve the node $v_i \in V$ corresponding to unlabeled samples on the target domain when $g_i=1$ and to remove the node $v_i$ when $g_i=0$.

\noindent\textbf{Prediction Dissimilarity based Edge Pruning (PDEP).} As stated above, when $a_{ij}=1$, the dissimilar prediction distributions between nodes of an edge in the graph might also lead to noisy connectivity in the graph, making a negative effect on semantic propagation. 
To remedy this, we first calculate the similarity score between the predicted label distributions of two nodes over an edge using dot product operation, and then threshold the similarity scores with a scalar $\kappa$ so as to obtain more credible graph connectivity. Therefore, we can formularize it as follows, 
\begin{equation}
    \label{graph_gij}
    \tilde{g}_{ij} = \neg a_{ij} \vee \mathds{1}\{\mathbf{p}_i^{\mathsf{T}}\mathbf{p}_j > \kappa\},
\end{equation}
where $\mathbf{p}_i=p(x_i)$ and $\mathbf{p}_j=p(x_j)$. Besides,  the binary indicator $\tilde{g}_{ij}$ serves to prune the graph edge connecting between nodes $v_i$ (i.e., unlabeled target node $x_i$) and $v_j$ (i.e., the sample $x_j$ from labeled source or target data) when $\tilde{g}_{ij}=0$; otherwise, the edge is preserved when $\tilde{g}_{ij}=1$. In this manner, PDEP can effectively control the removal and preservation of the graph edge only when $a_{ij}=1$. However, when $a_{ij}=0$, PDEP becomes invalid, ensuring that the edge connecting nodes $v_i$ and $v_j$ in the graph will be consistently preserved. 

Upon executing CUNR and PDEP, we can integrate Eq. (\textcolor{black}{\ref{graph_gij}}) into Eq. (\textcolor{black}{\ref{graph_gi1}}) and then revise Eq. (\textcolor{black}{\ref{graph_gi1}}) for the rebuilt graph connectivity as follows,
\begin{equation}
    \label{graph_gi2}
    g_i^{j} = g_i \cdot \tilde{g}_{ij}.
\end{equation}

\noindent
Here, the subscript $i$ of $g_i^{j}$ denotes the unlabeled target node $x_i$, and the superscript $j$ indicates the other node $x_j$ (from labeled source or target data) on the same edge as $x_i$.
 
Once we have the indicator $g_i^j$, we will be able to achieve more reliable connectivity between nodes in the graph $G$, hence improving the performance of semantic propagation.
 
\subsection{Adaptive Betweenness Clustering} 
In this section, we conduct semantic transfer for categorical domain alignment using the updated graph that represents the reliable structure of the training data. Here, we propose a newly devised clustering algorithm for semantic propagation, called Adaptive Betweenness Clustering (ABC). On the basis of the rebuilt graph, such an algorithm propagates and aggregates labels along graph edges. This allows the semantic label information of a labeled sample to be transferred to an unlabeled sample with clustering between samples. Using this approach in the spherical feature space provided by the prototypical classifier, the sample's predicted probability distributions indicate the cosine similarities between the feature and the prototypes for each category. Hence, this algorithm enforces a pair of samples with the same ground-truth (labeled) or predicted (unlabeled) class labels to have highly similar probability distributions. When the probability distribution of the latter is brought closer to that of the former, semantic propagation from labeled to unlabeled instances is achieved.
 
In specific, we first generate a sample pair from a labeled sample $x_i$ and an unlabeled sample $x_j$ by setting $s_{ij}=1$ as a pairwise label if $x_i$ and $x_j$ belong to the same class, otherwise $s_{ij}=0$ for different classes. Then, we adopt a binary cross-entropy loss to draw samples from the same class closer together in the feature space while separating samples from other classes. Utilizing a pairwise label as a target, adaptive betweenness clustering thus can be computed with the following loss:
\begin{equation}
    \label{Equation:BinaryCE}
     \pi(x_i, x_j, s_{ij})=-[s_{ij}\log(\mathbf{p}_i^{\mathsf{T}}\mathbf{p}_j)+(1-s_{ij})\log(1-\mathbf{p}_i^{\mathsf{T}}\mathbf{p}_j)],
\end{equation}
where $\mathbf{p}_i=p(x_i)$ and $\mathbf{p}_j=p(x_j)$. As demonstrated in~\cite{rebuffi2020lsd, li2021cross}, perturbations integrated into unlabeled target data can significantly improve the performance of the model; hence, we here augment unlabeled examples from the target domain for better propagation. 

Due to domain shift in SSDA, the semantic information of labeled source samples, though is large in volume, is less correlated with unlabeled target examples, whereas the target label information of labeled samples has a greater correlation with unlabeled samples on the target domain, but is relatively scarce.
Hence, to enable cross-domain semantic alignment, we propose across-domain betweenness clustering (ADBC) and within-domain betweenness clustering (WDBC) to propagate semantic label information from labeled instances on the source and target domains to unlabeled target data.
With the distinct but complimentary characteristics of semantic information from source and target labels, semantic transfer can be conducted with the following losses with regard to unlabeled target samples:
\begin{equation}
    \label{Equation:ABCLoss}
    {\mathcal{L}}^{abc}={\mathcal{L}}^{wdbc}+{\mathcal{L}}^{adbc},
\end{equation}
\begin{equation}
    \label{Equation:WdbcLoss}
    \mathcal{L}^{wdbc} = \frac{1}{N_u}  \sum_{i \in I} \frac{1}{N_l} \sum_{j \in P} {g_i^{j} \cdot \pi(x_i, x_{j}, a_{ij})},
\end{equation}
\begin{equation}
    \label{Equation:AdbcLoss}
    \mathcal{L}^{adbc} = \frac{1}{N_u} \sum_{i \in I} \frac{1}{N_s} \sum_{j^{\prime} \in Q} {g_i^{j^{\prime}} \cdot \pi(x_i, x_{j^{\prime}}, a_{ij^{\prime}})},
\end{equation}

where $I$, $P$ and $Q$ denote the collections of unlabeled samples from the target domain, and labeled ones from the target and source domains, respectively, whose sample indexes are denoted by $i\in\{1,2,\cdots, N_u\}$, $j\in\{1,2,\cdots, N_l\}$ and $j^{\prime}\in\{N_l+1,N_l+2,\cdots, N_l+N_s\}$.  

\begin{figure}[t]
    \centering
    \begin{minipage}{0.24\textwidth}
        \includegraphics[width=\linewidth, height=3.5cm]{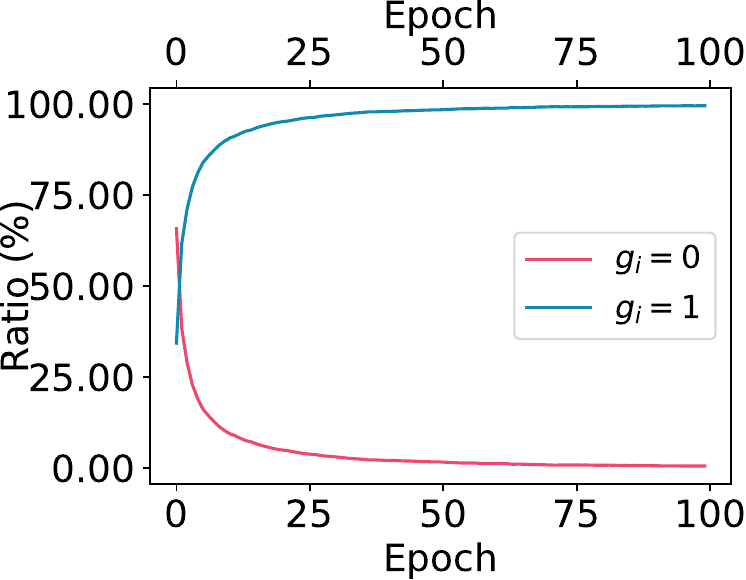}
        \caption*{\quad\quad(a)}
    \end{minipage}
    \hfill
    \begin{minipage}{0.24\textwidth}
        \includegraphics[width=\linewidth, height=3.5cm]{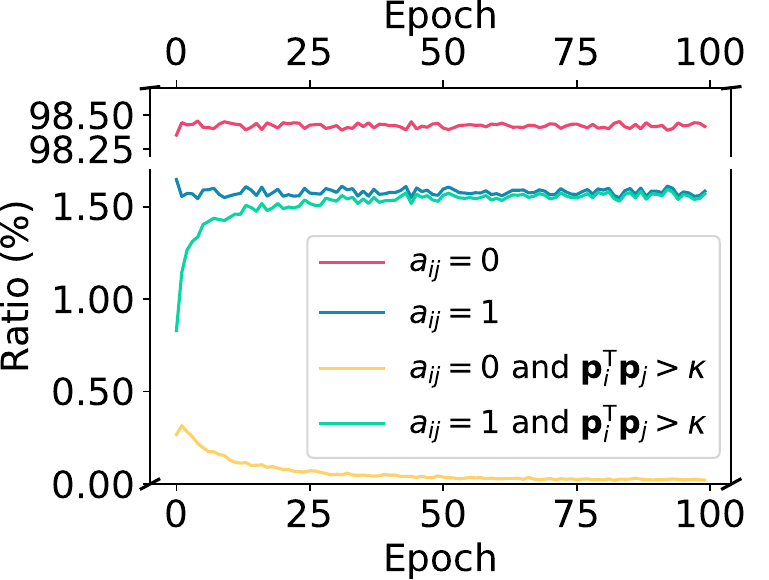}
        \caption*{\quad\quad(b)}
    \end{minipage}
    \caption{Empirical analysis on the proportion of unlabeled target samples that meet the conditions proposed in the CUNR and PDEP strategies during the model training process. (a) and (b) are specified in the conditions introduced by Eqs. (\textcolor{black}{\ref{graph_w}}) and (\textcolor{black}{\ref{graph_gi1}}). For (b), it is important to note that due to the large number of classes in the dataset, only a small fraction of the sample pairs randomly generated in each epoch consist of an unlabeled target sample and a labeled source or target sample with the same label. Therefore, the proportion of unlabeled target domain samples satisfying ``$a_{ij}=1$'' will be much smaller compared to the proportion of that satisfying ``$a_{ij}=0$''. The experiment is performed with ``R $\rightarrow$ P'' on Office-Home using ResNet-34, under the 3-shot setup.}
    \label{Figure:Ratio}
\end{figure}

Here, to gain a better understanding of how Eqs. (\textcolor{black}{\ref{graph_w}})-(\textcolor{black}{\ref{Equation:BinaryCE}}) work well in Eq. (\textcolor{black}{\ref{Equation:WdbcLoss}}) (Eq. (\textcolor{black}{\ref{Equation:AdbcLoss}}) follows the same rule as Eq. (\textcolor{black}{\ref{Equation:WdbcLoss}})), we provide the following detailed clarifications.

\begin{itemize}

    \item According to Eq.~(\textcolor{black}{\ref{graph_gi1}}), when $g_{i}>0$ (i.e. $g_i = 1$), the model prediction remains confident. At this time, if $a_{ij}=1$ (i.e. $s_{ij}=1$ in Eq. (\textcolor{black}{\ref{Equation:BinaryCE}})) and $\mathbf{p}_i^{\mathsf{T}}\mathbf{p}_j > \kappa$ (in Eq. (\textcolor{black}{\ref{graph_gij}})), then $g_i^j > 0$ (i.e. $g_i^j = 1$ in Eq. (\textcolor{black}{\ref{graph_gi2}})), and ``$s_{ij}\log(\mathbf{p}_i^{\mathsf{T}}\mathbf{p}_j)$'' (i.e., the first positive term of Eq. (\textcolor{black}{\ref{Equation:BinaryCE}})) contributes to the loss in Eq. (\textcolor{black}{\ref{Equation:WdbcLoss}}). In this case, $x_i$ and $x_j$ should be close to each other in the feature space.

    \item Still under the condition that  $g_i > 0$ (i.e. $g_i = 1$) in Eq.~(\textcolor{black}{\ref{graph_gi1}}), when $a_{ij}=0$, $\tilde{g}_{ij}=1$ in Eq. (\textcolor{black}{\ref{graph_gij}})) always holds true, so $g_i^j > 0$ (i.e. $g_i^j = 1$). In this case, ``$(1-s_{ij})\log(1-\mathbf{p}_i^{\mathsf{T}}\mathbf{p}_j)$'', namely the second negative term in Eq. (\textcolor{black}{\ref{Equation:BinaryCE}}), contributes to the loss of Eq. (\textcolor{black}{\ref{Equation:WdbcLoss}}), where $x_i$ and $x_j$ should be separated from each other. 
    
    \item Finally, when $g_{i}<0$ (i.e. $g_i = 0$), or when $a_{ij}=0$ and $\mathbf{p}_i^{\mathsf{T}}\mathbf{p}_j < \kappa$, then $g_j^i < 0$ (i.e. $g_i^j = 0$), and thus in Eq. (\textcolor{black}{\ref{Equation:BinaryCE}}), neither ``$s_{ij}\log(\mathbf{p}_i^{\mathsf{T}}\mathbf{p}_j)$'' nor ``$(1-s_{ij})\log(1-\mathbf{p}_i^{\mathsf{T}}\mathbf{p}_j)$'' contributes to the loss for Eq. (\textcolor{black}{\ref{Equation:WdbcLoss}}).

\end{itemize}
 
To provide a more intuitive demonstration, we conducted empirical analysis on the proportion of unlabeled target samples that meet the conditions specified in Eqs. (\textcolor{black}{\ref{graph_w}}) and (\textcolor{black}{\ref{graph_gi1}}) proposed in the CUNR and PDEP strategies during the model training process. The results are visualized in Fig.~\ref{Figure:Ratio}. Fig.~\ref{Figure:Ratio}(a) shows that as the model trains, the number of unlabeled samples satisfying the condition ``$\mathit{g}_{ij}=1$'' gradually increases, indicating an increasing level of predicted confidence in the unlabeled target samples. Moreover, in Fig.~\ref{Figure:Ratio}(b), it can be observed that as the training progresses, the ratio of unlabeled target domain samples satisfying the condition ``$\mathit{a}_{ij}=1$ and $\mathbf{p}_i^{\mathsf{T}}\mathbf{p}_j > \kappa$'' approaches the ratio of that satisfying the condition ``$\mathit{a}_{ij}=1$''. This demonstrates that the similarity between the prediction distributions of the unlabeled target samples and the labeled samples from the source or target domains increases gradually during the training process. On the contrary, the number of unlabeled target samples satisfying the condition ``$\mathit{a}_{ij}=0$ and $\mathbf{p}_i^{\mathsf{T}}\mathbf{p}_j > \kappa$'' decreases as the training progresses and may eventually approach zero. These findings support our assumption that when two samples, one from the unlabeled target samples and the other from the labeled samples of the source or target domain, do not belong to the same class, their prediction distributions will be dissimilar.

\subsection{Further Optimization of G-ABC based SSDA}

After achieving cross-domain semantic alignment, we apply auxiliary techniques, such as pseudo-label selection, self-training, and consistency training, to further enhance the model training.

\noindent
\textbf{Pseudo-label Selection.} Due to the scarcity of target labels, overfitting is likely to occur when the ${\mathcal{L}}^{wdbc}$ loss is applied. To mitigate this issue, we apply a pseudo-label selection strategy to unlabeled target samples and preserve pseudo-labels with high confidence to increase the number of target labels, hence enhancing the semantic label diversity on the target domain. In this work, pseudo-label selection employs the prediction capability of a model to generate artificial hard labels for a subset of unlabeled target samples; hence, a collection of pseudo-labeled target samples, denoted by $\mathcal{D}_{pu}$, can be obtained as follows,
\begin{equation}
    \label{dpu}
    \begin{split}
    \mathcal{D}_{pu} \leftarrow &\{(x,\hat{y})|{\hat{y}}=\mathop{{\argmax}_{k}}{(p(x)[k])}, \\
        &\max(p(x))>{\tau}^{\prime}, \forall x \in {\mathcal{D}_{u}}\},
    \end{split}
\end{equation}
where ${\tau}^{\prime}$ denotes another higher confidence threshold than $\tau$ in Eq. (\textcolor{black}{\ref{graph_gi1}}). Noted that $\mathcal{D}_{pu}$ is only used in the ${\mathcal{L}}^{wdbc}$ loss, namely Eq. (\textcolor{black}{\ref{Equation:WdbcLoss}}).

\noindent
\textbf{Self-training.} According to~\cite{mishra2021surprisingly, sohn2020fixmatch}, we adopt self-training to boost the model's robustness against the selected samples in $\mathcal{D}_{pu}$. Specifically, we employ the progressive self-training technique described in~\cite{sohn2020fixmatch}, termed label consistency, in which the model is constrained to generate the same output when the selected images are augmented with slight perturbations.
In practice, label consistency can be implemented through the following loss:
\begin{equation}
    \label{Equation:LabConsLoss}
    \mathcal{L}^{lab} =-\frac{1}{N_{pu}}\sum_{(x_i, y_i)\in{\mathcal{D}_{pu}}}p_y(y_i)\log(p(\textbf{Aug}(x_i))),
\end{equation}
where $N_{pu}=|\mathcal{D}_{pu}|$ indicates the sample size of $\mathcal{D}_{pu}$,  $p_y(\cdot)$ represents the function to create a one-hot probability vector for a pseudo-label,  and $\textbf{Aug}(\cdot)$ denotes the function to perturb the input images. 

\noindent
\textbf{Consistency Training.}
The pseudo-label selection mechanism fails to assign pseudo-labels to all unlabeled target samples. In accordance with~\cite{xie2019unsupervised}, we can adequately leverage unlabeled target samples through consistency training, hence increasing the smoothness of the model.
According to~\cite{xie2019unsupervised}, we can achieve this by preserving the consistency of the model's output distributions between unlabeled target samples and their perturbed counterparts using Kullback–Leibler (KL) divergence as follows,
\begin{equation}
    \label{Equation:ConsTrain}
    \mathcal{L}^{con}=\frac{1}{N_u}\sum_{x_i\in{\mathcal{D}_{u}}} \tilde{p}(x_i)\log(\frac{\tilde{p}(x_i)}{p(\textbf{Aug}(x_i))}),
\end{equation}
\begin{equation}
    \tilde{p}(x_i)=\Sharpen(p(x_i))=\frac{(p(x_i))^{\frac{1}{T^{\prime}}}}{\sum_{k=1}^K{({p(x_i)}_{k})^{\frac{1}{T^{\prime}}}}},
    \label{Equation:Sharpening}
\end{equation}
where $k$ indicates the $k$-th element of the target distribution vector $p(x_{i}^u)$ and  $T^{\prime}$ is the temperature factor. It should be noticed that different from~\cite{xie2019unsupervised}, we here use a ``soft" distribution with a sharpening function $\Sharpen(\cdot)$ proposed in~\cite{li2020dividemix, berthelot2019mixmatch} to sharpen the observed probability distribution $\tilde{p}(x_i)$, thereby driving the model to generate lower-entropy predictions.
 
\begin{table*}
    \centering
    \footnotesize
    \caption{Results (\%) on {DomainNet} under the settings of 1-shot and 3-shot with both AlexNet (ANet) and ResNet-34(RN-34) backbones. The top best methods are in \textbf{bold}. (Mean accuracy and standard variance over 3 trials)}
    \resizebox{\textwidth}{!}{
    \begin{tabular}{c|l|cccccccccccccc|cc}
    \toprule[1.5pt]
    \toprule
    \multirow{2}[2]{*}{Net} & \multicolumn{1}{c|}{\multirow{2}[2]{*}{Method}} & \multicolumn{2}{c}{R→C} & \multicolumn{2}{c}{R→P} & \multicolumn{2}{c}{P→C} & \multicolumn{2}{c}{C→S} & \multicolumn{2}{c}{S→P} & \multicolumn{2}{c}{R→S} & \multicolumn{2}{c|}{P→R} & \multicolumn{2}{c}{Mean} \\
    &       & 1-shot & 3-shot & 1-shot & 3-shot & 1-shot & 3-shot & 1-shot & 3-shot & 1-shot & 3-shot & 1-shot & 3-shot & 1-shot & 3-shot & 1-shot & 3-shot \\
    \midrule
    \multirow{13}[2]{*}{ANet} & S+T~\cite{saito2019semi}   & 43.3  & 47.1  & 42.4  & 45.0  & 40.1  & 44.9  & 33.6  & 36.4  & 35.7  & 38.4  & 29.1  & 33.3  & 55.8  & 58.7  & 40.0  & 43.4 \\
    & DANN~\cite{saito2019semi}  & 43.3  & 46.1  & 41.6  & 43.8  & 39.1  & 41.0  & 35.9  & 36.5  & 36.9  & 38.9  & 32.5  & 33.4  & 53.5  & 57.3  & 40.4  & 42.4 \\
    & MME~\cite{saito2019semi}   & 48.9  & 55.6  & 48.0  & 49.0  & 46.7  & 51.7  & 36.3  & 39.4  & 39.4  & 43.0  & 33.3  & 37.9  & 56.8  & 60.7  & 44.2  & 48.2 \\
    & Meta-MME~\cite{li2020online} & -     & 56.4  & -     & 50.2  & -     & 51.9  & -     & 39.6  & -     & 43.7  & -     & 38.7  & -     & 60.7  & -     & 48.8 \\
    & BiAT~\cite{2020Bidirectional}  & 54.2  & 58.6  & 49.2  & 50.6  & 44.0  & 52.0  & 37.7  & 41.9  & 39.6  & 42.1  & 37.2  & 42.0  & 56.9  & 58.8  & 45.5  & 49.4 \\
    & APE~\cite{kim2020attract}   & 47.7  & 54.6  & 49.0  & 50.5  & 46.9  & 52.1  & 38.5  & 42.6  & 38.5  & 42.2  & 33.8  & 38.7  & {57.5} & 61.4  & 44.6  & 48.9 \\
    & PAC~\cite{mishra2021surprisingly}   & 55.4  & 61.7  & 54.6  & 56.9  & 47.0  & 59.8  & 46.9  & 52.9  & 38.6  & 43.9  & 38.7  & 48.2  & 56.7  & 59.7  & 48.3  & 54.7 \\
    & Relaxd-cGAN~\cite{luo2021relaxed} & -     & 56.8  & -     & 51.8  & -     & 52.0  & -     & 44.1  & -     & 44.2  & -     & 42.8  & -     & 61.1  & -     & 50.5 \\
    & ECACL-T~\cite{li2021ecacl} & 56.8  & 62.9  & 54.8  & 58.9  & \textbf{56.3} & 60.5  & 46.6  & 51.0  & \textbf{54.6} & 51.2  & 45.4  & \textbf{48.9} & 62.8  & 67.4  & 53.4  & 57.7 \\
    & ECACL-P~\cite{li2021ecacl} & 55.8  & 62.6  & 54.0  & 59.0  & 56.1  & 60.5  & 46.1  & 50.6  & \textbf{54.6} & 50.3  & 45.0  & 48.4  & 62.3  & 67.4  & 52.8  & 57.6 \\
    & ${\text{S}}^{\text{3}}{\text{D}}$~\cite{R3C2} &  53.5  &  56.5  &  51.8  &  52.2  &  49.1  &  53.9  &  40.1  &  44.4  &  44.9  &  48.7  &  39.9  &  39.2  &  61.7  &  65.4  &  48.7  &  51.5  \\
    & CLDA~\cite{singh2021clda}  & 56.3  & 59.9  & 56.0  & 57.2  & 50.8  & 54.6  & 42.5  & 47.3  & 46.8  & 51.4  & 38.0  & 42.7  & 64.4  & 67.0  & 50.7  & 54.3 \\
    & CDAC~\cite{li2021cross}  & 56.9  & 61.4  & 55.9  & 57.5  & 51.6  & 58.9  & 44.8  & 50.7  & 48.1  & 51.7  & 44.1  & 46.7  & 63.8  & 66.8  & 52.1  & 56.2 \\
    & G-ABC (Ours) & \textbf{60.17$\pm$0.78} & \textbf{63.89$\pm$0.41} & \textbf{57.44$\pm$0.66} & \textbf{59.73$\pm$0.33} & 55.98$\pm$0.93 & \textbf{64.03$\pm$0.34} & \textbf{48.75$\pm$0.50} & \textbf{53.42$\pm$0.57} & 54.11$\pm$0.61 & \textbf{56.36$\pm$0.54} & \textbf{47.09$\pm$0.91} & 48.17$\pm$0.35 & \textbf{67.84$\pm$0.79} & \textbf{70.78$\pm$0.36} & \textbf{55.92} & \textbf{59.68} \\
    \midrule
    \multirow{15}[2]{*}{RN-34} & S+T~\cite{saito2019semi}   & 55.6  & 60.0  & 60.6  & 62.2  & 56.8  & 59.4  & 50.8  & 55.0  & 56.0  & 59.5  & 46.3  & 50.1  & 71.8  & 73.9  & 56.9  & 60.0 \\
    & DANN~\cite{saito2019semi}  & 58.2  & 59.8  & 61.4  & 62.8  & 56.3  & 59.6  & 52.8  & 55.4  & 57.4  & 59.9  & 52.2  & 54.9  & 70.3  & 72.2  & 58.4  & 60.7 \\
    & MME~\cite{saito2019semi}   & 70.0  & 72.2  & 67.7  & 69.7  & 69.0  & 71.7  & 56.3  & 61.8  & 64.8  & 66.8  & 61.0  & 61.9  & 76.1  & 78.5  & 66.4  & 68.9 \\
    & UODA~\cite{qin2021contradictory}  & 72.7  & 75.4  & 70.3  & 71.5  & 69.8  & 73.2  & 60.5  & 64.1  & 66.4  & 69.4  & 62.7  & 64.2  & 77.3  & 80.8  & 68.5  & 71.2 \\
    & Meta-MME~\cite{li2020online} & -     & 73.5  & -     & 70.3  & -     & 72.8  & -     & 62.8  & -     & 68.0  & -     & 63.8  & -     & 79.2  & -     & 70.1 \\
    & BiAT~\cite{2020Bidirectional}  & 73.0  & 74.9  & 68.0  & 68.8  & 71.6  & 74.6  & 57.9  & 61.5  & 63.9  & 67.5  & 58.5  & 62.1  & 77.0  & 78.6  & 67.1  & 69.7 \\
    & APE~\cite{kim2020attract}   & 70.4  & 76.6  & 70.8  & 72.1  & 72.9  & 76.7  & 56.7  & 63.1  & 64.5  & 66.1  & 63.0  & 67.8  & 76.6  & 79.4  & 67.6  & 71.7 \\
    & ELP~\cite{ELP}   & 72.8  & 74.9  & 70.8  & 72.1  & 72.0  & 74.4  & 59.6  & 63.3  & 66.7  & 69.7  & 63.3  & 64.9  & 77.8  & 81.0  & 69.0  & 71.6 \\
    & PAC~\cite{mishra2021surprisingly}   & 74.9  & 78.6  & 73.0  & 74.3  & 72.6  & 76.0  & 65.8  & 69.6  & 67.9  & 69.4  & 68.7  & 70.2  & 76.7  & 79.3  & 71.4  & 73.9 \\
    & DECOTA\cite{YangLuyu2020DCwT} & 79.1  & 80.4  & 74.9  & 75.2  & 76.9  & 78.7  & 65.1  & 68.6  & 72.0  & 72.7  & 69.7  & 71.9  & 79.6  & 81.5  & 73.9  & 75.6 \\
    & ECACL-T~\cite{li2021ecacl} & 73.5  & 76.4  & 72.8  & 74.3  & 72.8  & 75.9  & 65.1  & 65.3  & 70.3  & 72.2  & 64.8  & 68.6  & 78.3  & 79.7  & 71.1  & 73.2 \\
    & ECACL-P~\cite{li2021ecacl} & 75.3  & 79.0  & 74.1  & 77.3  & 75.3  & 79.4  & 65.0  & 70.6  & 72.1  & 74.6  & 68.1  & 71.6  & 79.7  & 82.4  & 72.8  & 76.4 \\
    & ${\text{S}}^{\text{3}}{\text{D}}$~\cite{R3C2} & 73.3 & 75.9  & 68.9  & 72.1  & 73.4  & 75.1  & 60.8  & 64.4  & 68.2  & 70.0  & 65.1  & 66.7  & 79.5  & 80.3  & 69.9  & 72.1 \\
    & UODAv2~\cite{R3C1}  & 77.0  & 79.4  & 75.4  & 76.7  & 75.5  & 78.3  & 66.5  & 70.2  & 72.1  & 74.2  & 70.9  & 72.1  & 79.7  & 82.3  & 73.9  & 76.2  \\
    & MCL~\cite{R3C3} & 77.4 & 79.4 & 74.6 & 76.3 & 75.5 & 78.8 & 66.4 & 70.9 & 74.0 & 74.7 & 70.7 & 72.3 & 82.0 & 83.3 & 74.4 & 76.5 \\
    & CLDA~\cite{singh2021clda}  & 76.1  & 77.7  & 75.1  & 75.7  & 71.0  & 76.4  & 63.7  & 69.7  & 70.2  & 73.7  & 67.1  & 71.1  & 80.1  & 82.9  & 71.9  & 75.3 \\
    & CDAC~\cite{li2021cross}  & 77.4  & 79.6  & 74.2  & 75.1  & 75.5  & 79.3  & 67.6  & 69.9  & 71.0  & 73.4  & 69.2  & 72.5  & 80.4  & 81.9  & 73.6  & 76.0 \\
    & G-ABC (Ours) & \textbf{80.74$\pm$0.41} & \textbf{82.07$\pm$0.21} & \textbf{76.84$\pm$0.63} & \textbf{76.72$\pm$0.32} & \textbf{79.26$\pm$0.19} & \textbf{81.57$\pm$0.41} & \textbf{71.95$\pm$0.47} & \textbf{73.68$\pm$0.38} & \textbf{75.04$\pm$0.54} & \textbf{76.27$\pm$0.20} & \textbf{73.21$\pm$0.32} & \textbf{74.28$\pm$0.09} & \textbf{83.42$\pm$0.61} & \textbf{83.87$\pm$0.28} & \textbf{77.47} & \textbf{78.23} \\
    \bottomrule
    \bottomrule[1.5pt]
    \end{tabular}%
    }
    \label{Table:DomainNet}
\end{table*}

\noindent
\textit{\textbf{Training Objectives.}}The loss function for optimizing the model can be expressed as a combination of the cross-entropy loss $\mathcal{L}^{ce}$ over all labeled samples accessible across domains and additional losses as previously described, i.e.,
\begin{equation}
    \label{Equation:OverallLoss}
    \mathcal{L}^{Overall}=\mathcal{L}^{ce}+\mathcal{L}^{lab}+\alpha\mathcal{L}^{con}+\beta\mathcal{L}^{abc},
\end{equation}
where $\alpha$ and $\beta$ are scalar hyper-parameters for loss weights. The model is updated using stochastic gradient descent (SGD) for backpropagation. 

\section{Experiments} 
\label{Section:Experiment}

\subsection{Datasets}
The proposed G-ABC approach is evaluated on three widely used benchmark datasets, including {\bf DomainNet}~\cite{peng2019moment}, {\bf Office-Home}~\cite{venkateswara2017deep} and {\bf Office-31}~\cite{saenko2010adapting}. To conduct a fair comparison, we follow all configurations of adaptation scenarios on different datasets as considered in~\cite{saito2019semi, qin2021contradictory, kim2020attract, li2021cross}, while each category has one-shot or three-shot samples with labels available in the target domain during training.
 
\noindent
\textbf{DomainNet}, consisting of 345 classes and six domains, is a large-scale benchmark dataset designed to evaluate multi-source domain adaption approaches. Following~\cite{saito2019semi}, we use a subset of the dataset proposed in~\cite{peng2019moment} as one of our evaluation benchmarks. Similar to MME~\cite{saito2019semi}, we only chose 4 domains: Real ({\bf R}), Clipart ({\bf C}), Painting ({\bf P}), and Sketch ({\bf S}) (each comprises 126 categories of images), as other domains and categories may contain samples with excessive noise. In accordance with~\cite{saito2019semi, qin2021contradictory, kim2020attract, li2021cross}, we construct experiments on seven adaptation scenarios employing these four domains.
 
\noindent
\textbf{Office-Home} is a notable SSDA benchmark dataset containing numerous challenging adaptation scenarios. There are 65 classes in this collection, and the available domains are Real ({\bf R}), Clipart ({\bf C}), Art ({\bf A}), and Product ({\bf P}). To achieve a fair comparison, we apply 12 adaptation scenarios to this dataset compared to previous SSDA methods, including~\cite{saito2019semi, qin2021contradictory, kim2020attract, li2021cross, YangLuyu2020DCwT}.
 
\begin{table*}
    \centering
    \scriptsize
    \caption{Results (\%) on {Office-Home} under the setting of 3-shot  with both AlexNet (ANet) and ResNet-34 (RN-34) backbones. The top best methods are in \textbf{bold}. (Mean accuracy and standard variance over 3 trials)}
    \resizebox{\textwidth}{!}{
    \begin{tabular}{c|p{7.565em}|cccccccccccc|c}
    \toprule[1.5pt]
    \toprule
    Net   & \multicolumn{1}{l|}{Method} & R→C   & R→P   & R→A   & P→R   & P→C   & P→A   & A→P   & A→C   & A→R   & C→R   & C→A   & C→P   & Mean \\
    \midrule
    \multirow{10}[2]{*}{ANet} & \multicolumn{1}{l|}{S+T~\cite{saito2019semi}} & 44.6  & 66.7  & 47.7  & 57.8  & 44.4  & 36.1  & 57.6  & 38.8  & 57    & 54.3  & 37.5  & 57.9  & 50.0 \\
          & \multicolumn{1}{l|}{DANN~\cite{saito2019semi}} & 47.2  & 66.7  & 46.6  & 58.1  & 44.4  & 36.1  & 57.2  & 39.8  & 56.6  & 54.3  & 38.6  & 57.9  & 50.3 \\
          & \multicolumn{1}{l|}{MME~\cite{saito2019semi}} & 51.2  & 73.0    & 50.3  & 61.6  & 47.2  & 40.7  & 63.9  & 43.8  & 61.4  & 59.9  & 44.7  & 64.7  & 55.2 \\
          & \multicolumn{1}{l|}{Meta-MME~\cite{li2020online}} & 50.3  & -     & -     & -     & 48.3  & 40.3  & -     & 44.5  & -     & -     & 44.5  & -     & - \\
          & \multicolumn{1}{l|}{BiAT~\cite{2020Bidirectional}} & -     & -     & -     & -     & -     & -     & -     & -     & -     & -     & -     & -     & 56.4 \\
          & \multicolumn{1}{l|}{APE~\cite{kim2020attract}} & 51.9  & 74.6  & 51.2  & 61.6  & 47.9  & 42.1  & 65.5  & 44.5  & 60.9  & 58.1  & 44.3  & 64.8  & 55.6 \\
          & \multicolumn{1}{l|}{PAC~\cite{mishra2021surprisingly}} & \textbf{58.9} & 72.4  & 47.5  & 61.9  & \textbf{53.2} & 39.6  & 63.8  & \textbf{49.9} & 60.0    & 54.5  & 36.3  & 64.8  & 55.2 \\
          & \multicolumn{1}{l|}{CLDA~\cite{singh2021clda} } & 51.5  & 74.1  & \textbf{54.3} & \textbf{67} & 47.9  & 47    & 65.8  & 47.4  & \textbf{66.6} & \textbf{64.1} & \textbf{46.8} & 67.5  & \textbf{58.3} \\
          & \multicolumn{1}{l|}{CDAC~\cite{li2021cross}} & 54.9  & 75.8  & 51.8  & 64.3  & 51.3  & \textbf{43.6} & 65.1  & 47.5  & 63.1  & 63.0    & 44.9  & 65.6  & 56.8 \\
            & G-ABC (Ours) & 55.12$\pm$0.71 & \textbf{76.21$\pm$0.59} & 53.20$\pm$0.45 & 64.59$\pm$0.43 & 50.45$\pm$0.70 & 41.76$\pm$0.51 & \textbf{67.41$\pm$0.67} & 47.51$\pm$0.87 & 62.07$\pm$0.93 & 63.52$\pm$0.97 & 42.72$\pm$0.81 & \textbf{68.23$\pm$0.46} & 57.73 \\
    \midrule
    \multirow{10}[2]{*}{RN-34} & \multicolumn{1}{l|}{S+T~\cite{saito2019semi}} & 55.7  & 80.8  & 67.8  & 73.1  & 53.8  & 63.5  & 73.1  & 54.0    & 74.2  & 68.3  & 57.6  & 72.3  & 66.2 \\
          & \multicolumn{1}{l|}{DANN~\cite{saito2019semi}} & 57.3  & 75.5  & 65.2  & 69.2  & 51.8  & 56.6  & 68.3  & 54.7  & 73.8  & 67.1  & 55.1  & 67.5  & 63.5 \\
          & \multicolumn{1}{l|}{MME~\cite{saito2019semi}} & 64.6  & 85.5  & 71.3  & 80.1  & 64.6  & 65.5  & 79    & 63.6  & 79.7  & 76.6  & 67.2  & 79.3  & 73.1 \\
          & \multicolumn{1}{l|}{Meta-MME~\cite{li2020online}} & 65.2  & -     & -     & -     & 64.5  & 66.7  & -     & 63.3  & -     & -     & 67.5  & -     & - \\
          & \multicolumn{1}{l|}{APE~\cite{kim2020attract}} & 66.4  & 86.2  & 73.4  & 82.0    & 65.2  & 66.1  & 81.1  & 63.9  & 80.2  & 76.8  & 66.6  & 79.9  & 74.0 \\
          & \multicolumn{1}{l|}{Relaxed-cGAN~\cite{luo2021relaxed}} & 68.4  & 85.5  & 73.8  & 81.2  & 68.1  & 67.9  & 80.1  & 64.3  & 80.1  & 77.5  & 66.3  & 78.3  & 74.2 \\
          & \multicolumn{1}{l|}{DECOTA~\cite{YangLuyu2020DCwT}} & \textbf{70.4} & 87.7  & 74.0    & 82.1  & 68.0    & 69.9  & 81.8  & 64    & 80.5  & 79    & 68.0    & 83.2  & 75.7 \\
          & \multicolumn{1}{l|}{CLDA~\cite{singh2021clda} } & 66.0    & 87.6  & \textbf{76.7} & 82.2  & 63.9  & \textbf{72.4} & 81.4  & 63.4  & \textbf{81.3} & 80.3  & \textbf{70.5} & 80.9  & 75.5 \\
          & \multicolumn{1}{l|}{CDAC~\cite{li2021cross}} & 67.8  & 85.6  & 72.2  & 81.9  & 67    & 67.5  & 80.3  & 65.9  & 80.6  & 80.2  & 67.4  & 81.4  & 74.2 \\
        & G-ABC (Ours) & 70.02$\pm$0.18 & \textbf{88.09$\pm$0.27} & 75.96$\pm$0.48 & \textbf{82.81$\pm$0.11} & \textbf{69.27$\pm$0.53} & 70.54$\pm$0.42 & \textbf{83.78$\pm$0.31} & \textbf{67.24$\pm$0.14} & 80.37$\pm$0.10 & \textbf{80.18$\pm$0.44} & 69.22$\pm$0.25 & \textbf{83.89$\pm$0.62} & \textbf{77.19} \\
    \bottomrule
    \bottomrule[1.5pt]
    \end{tabular}%
    }
    \label{Table:OfficeHome}
\end{table*}

\begin{table*}[htbp]
\scriptsize
  \centering
    \caption{Results (\%) on Office-31 under the settings of 1-shot and 3-shot with the AlexNet backbone.  The top best methods are in \textbf{bold}. (Mean accuracy and standard variance over 3 trials)}
    \begin{tabular}{p{2.5cm} | p{1.2cm}<{\centering} p{1.2cm}<{\centering} p{1.2cm}<{\centering} p{1.2cm}<{\centering} | p{1.2cm}<{\centering} p{1.2cm}<{\centering}}
    \toprule[1.5pt]
    \toprule
    \multirow{2}[2]{*}{Method} & \multicolumn{2}{c}{W$\rightarrow$A} & \multicolumn{2}{c|}{D$\rightarrow$A} & \multicolumn{2}{c}{Mean} \\
          & 1-shot & 3-shot & 1-shot & 3-shot & 1-shot & 3-shot \\
    \midrule
    S+T~\cite{saito2019semi}   & 50.4  & 61.2  & 50.0  & 62.4  & 50.2  & 61.8  \\
    DANN~\cite{saito2019semi}  & 57.0  & 64.4  & 54.5  & 65.2  & 55.8  & 64.8  \\
    MME~\cite{saito2019semi}   & 57.2  & 67.3  & 55.8  & 67.8  & 56.5  & 67.6  \\
    BiAT~\cite{2020Bidirectional}  & 57.9  & 68.2  & 54.6  & 68.5  & 56.3  & 68.4  \\
    APE~\cite{kim2020attract}   & -     & 67.6  & -     & 69.0  & -     & 68.3  \\
    PAC~\cite{mishra2021surprisingly}   & 53.6  & 65.1  & 54.7  & 66.3  & 54.2  & 65.7  \\
    CLDA~\cite{singh2021clda} & {64.6} & {70.5} & 62.7 & {72.5} & {63.6} & {71.5} \\
    CDAC~\cite{li2021cross}  & 63.4  & 70.1  & {62.8}  & 70.0  & 63.1  & 70.0  \\
    G-ABC (Ours) & \textbf{67.9$\pm$1.26} & \textbf{70.97$\pm$0.48} & \textbf{65.73$\pm$1.03} & \textbf{73.06$\pm$0.35} & \textbf{66.81} & \textbf{72.02} \\
    \bottomrule
    \bottomrule[1.5pt]
    \end{tabular}%
  \label{Table:Office31}%
\end{table*}%

\noindent
\textbf{Office-31} consists of 31 object categories organized into three domains: Amazon ({\bf A}), DSLR ({\bf D}), and Webcam ({\bf W}). These categories contain objects frequently seen in offices, such as keyboards, file cabinets, and laptops. Following previous SSDA efforts~\cite{saito2019semi, qin2021contradictory, kim2020attract}, we select Amazon as the source domain since only Amazon is a large domain with sufficient samples for each class, whereas Webcam and DSLR do not. Therefore, we only consider two adaptation scenarios on this relatively smaller SSDA dataset benchmark, i.e., ``W $\rightarrow$ A'' and ``D $\rightarrow$ A''.

\subsection{Implementation}
\label{SubSection:Implementation}
To be fair, we adhere to the conventional SSDA task configurations from earlier research~\cite{saito2019semi, qin2021contradictory, kim2020attract}. Specifically, we first select AlexNet~\cite{2012ImageNet} and ResNet-34~\cite{he2016deep} with pre-trained weights on ImageNet~\cite{2012ImageNet} as the backbone networks for all our experiments. However, the last layer of each backbone is replaced with a prototypical classifier based on cosine similarity, followed by an unbiased linear neural network that takes normalized features from the feature extractor as inputs. Here, we optimize the entire model using mini-batch stochastic gradient descent (SGD) with momentum. In addition, throughout each iteration, we first train the model under supervision on all labeled samples from both domains to generate representative prototypes of each class, followed by the proposed ADBC and WDBC stages to further enhance the model. Moreover, we use RandAugment~\cite{2020Randaugment} and Cutout~\cite{devries2017improved} to generate perturbations for unlabeled target data used in Eq. (\textcolor{black}{\ref{Equation:WdbcLoss}}),  (\textcolor{black}{\ref{Equation:AdbcLoss}}), (\textcolor{black}{\ref{Equation:LabConsLoss}}), and (\textcolor{black}{\ref{Equation:ConsTrain}}). 

For fair comparisons, the majority of the remaining experimental settings in our proposed method are identical to previous SSDA efforts like~\cite{saito2019semi, qin2021contradictory, kim2020attract}. Similar to~\cite{saito2019semi, kim2020attract}, we implement all experiments on the PyTorch\footnote{https://pytorch.org/} platform. Besides, during each iteration, we randomly select four mini-batches from $\mathcal{D}_s$, $\mathcal{D}_l$, $\mathcal{D}_{pu}$, and $\mathcal{D}_{u}$, with batch sizes of 32, 32, 32, and 64 for AlexNet or 24, 24, 24, and 48 for ResNet-34. In addition, we employ the same learning rate schedule as~\cite{2014Unsupervised}, with the learning rate $\xi_{t}$ at the $t$-th iteration set as follows:
\begin{equation}
\xi_{t}=\frac{\xi_{0}}{{(1+0.0001\times{t})^{0.75}}},
\end{equation}
where $\xi_{0}$ represents the initial learning rate. To balance numerous loss terms, we set $\alpha$ and $\beta$ in Eq. (\textcolor{black}{\ref{Equation:OverallLoss}}) to 0.03 and 25.0. Then, based on~\cite{sohn2020fixmatch, li2021cross}, we set the confidence threshold to ${\tau}=0.95$ and $\tau^{\prime}=0.975$. In addition, we set the similarity threshold $\kappa$ to $0.20$. The value of temperatures involved in the construction of the model architecture and the sharpening function in Eq.~(\textcolor{black}{\ref{Equation:Sharpening}}) are set to 0.05 and 0.85, respectively. Due to the distinctness of each dataset and adaptation scenario, we set the total number of training epochs $\mathcal{T}$ to varying values. Note that $\mathcal{T}=100$ is a common value setting for a variety of application scenarios.

To choose the hyper-parameters, such as $\alpha$, $\beta$, $\tau$ and $\kappa$,  similar to \textbf{MME}~\cite{saito2019semi}, we selected three labeled examples as the validation set for the target domain and utilized these validation examples to choose the value choice of these hyper-parameters when the validation accuracy was at its highest. Also, during this process, we froze the other hyper-parameters while conducting experiments with a specific one.

\noindent
\textbf{Class-wise Similarity Score.} To assess the effectiveness of Adaptive Betweenness Clustering, we define a Class-wise Similarity Score (CSS) to measure the average prediction similarity between two classes, where one class $c$ originates from the unlabeled target domain and the other class $c^{\prime}$ comes from labeled source and target domains. In this way, we use $s(c, c^{\prime})$ to define the CSS between the class $c$ and $c^{\prime}$, and more specifically, $s(c, c^{\prime})$ can be formulated as follows,
\begin{equation}
    \label{Equation:SccScore}
    s(c, c^{\prime}) = \frac{1}{N_u^{c}} \sum_{i \in I^c} \frac{1}{N_l^{c^{\prime}}+N_s^{c^{\prime}}} \sum_{j \in P^{c\prime} \cup Q^{c\prime} } {\mathbf{p}_i^{\mathsf{T}}\mathbf{p}_j},
\end{equation}
where $I^c$, $P^{c^{\prime}}$ and $Q^{c^{\prime}}$ denote the collections of unlabeled target samples of class $c$, labeled target samples of the class ${c^{\prime}}$, and labeled source samples of class ${c^{\prime}}$, respectively, each containing instances with sizes of $N_u^{c}$, $N_l^{{c^{\prime}}}$ and $N_s^{{c^{\prime}}}$. In general, the larger the average prediction similarity between classes $c$ and $c^{\prime}$ is, the greater the CSS $s(c, c^{\prime})$ is. At this point, a higher CSS score indicates that there is a greater similarity in predictions between unlabeled samples from class ${c}$ in the target domain and the labeled source or target data from class ${c^{\prime}}$. This implicitly suggests that these samples are close to each other in the feature space.

\begin{figure*}[h]
    \centering
    \begin{minipage}{\textwidth}
        \centering
        \includegraphics[width=0.70\textwidth]{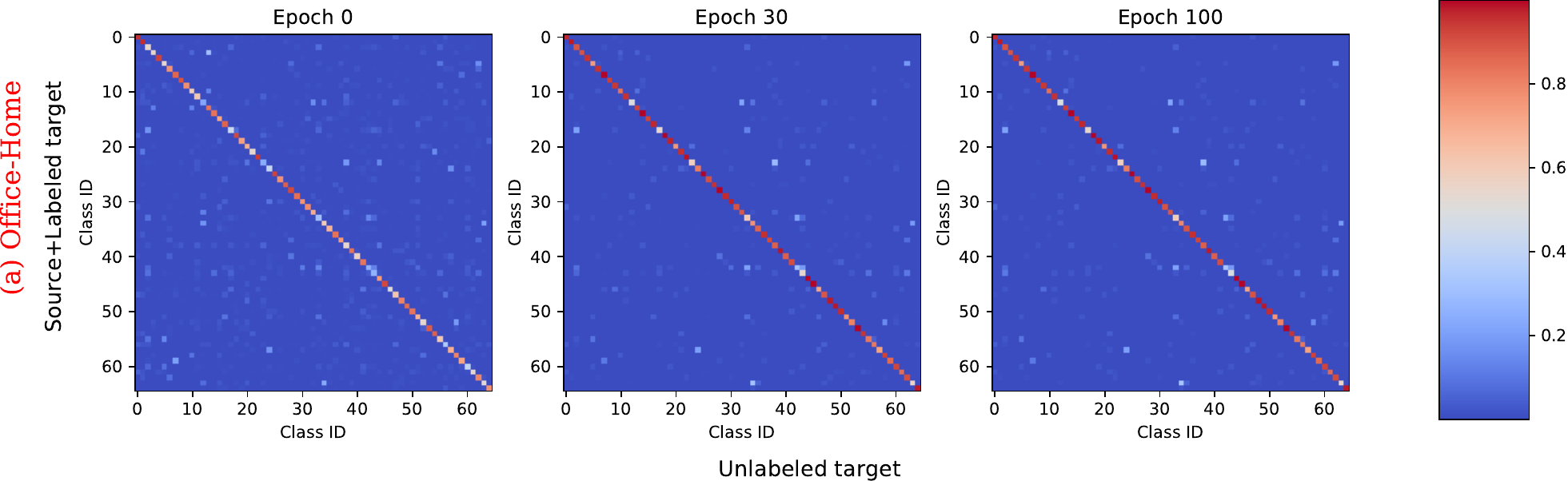}
    \end{minipage}
    \hfill
    \hfill
    \begin{minipage}{\textwidth}
        \centering
        \includegraphics[width=0.70\textwidth]{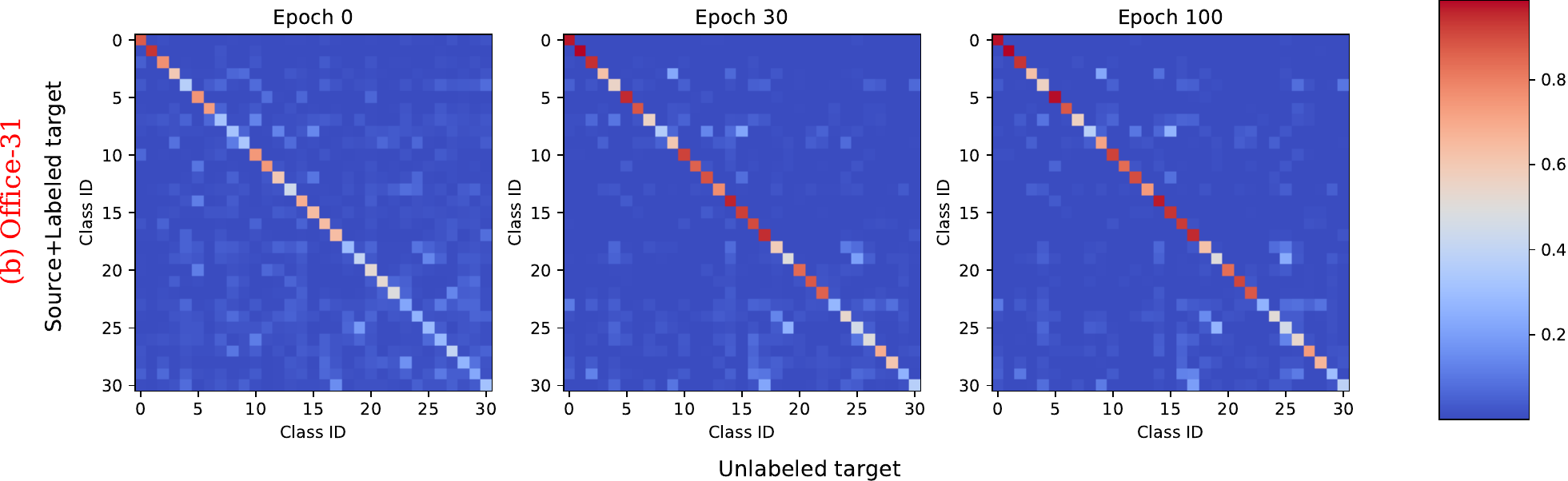}
    \end{minipage}
    \caption{The evaluation of Adaptive Betweenness Clustering involves analyzing the confusion matrices for each epoch on every dataset. Each element in these matrices is associated with a Class-wise Similarity Score, denoted as $s(c, c^{\prime})$. This score, defined by Eq.~(\textcolor{black}{\ref{Equation:SccScore}}) in Sec.~\ref{SubSection:Implementation}, quantifies the similarity between two classes,  $c$ and $c^{\prime}$. In this context, class $c$ refers to the class whose samples are the unlabeled target data, while class $c^{\prime}$ includes classes from both the labeled source and target domains. A higher $s(c, c^{\prime})$ score in each element suggests a greater similarity in predictions between the unlabeled samples from class $c$ in the target domain and the labeled source or target data from class $c^{\prime}$. The experiments are performed with  (a): ``R $\rightarrow$ P'' on Office-Home using ResNet-34, and (b): ``D$\rightarrow$A'' on Office-31 using AlexNet, respectively,  both under the 3-shot setup.}
    \label{Figure:Confusion}
\end{figure*}

\begin{table*}[t]
\scriptsize
  \centering
    \caption{Results (\%) of ablation study on DomainNet under the setting of 3-shot with the ResNet-34 backbone. $\dagger$ denotes that the adaptation model is trained by $\mathcal{L}^{adbc}$ or $\mathcal{L}^{wdbc}$ loss without incorporating perturbations into the samples. $\ddagger$ denotes that the pseudo-labeled target samples are not used in the $\mathcal{L}^{wdbc}$ loss. Moreover, $\clubsuit$ represents the removal of the first positive term, namely ``$s_{ij}\log(\mathbf{p}_i^{\mathsf{T}}\mathbf{p}_j)$'', from Eq.~(\ref{Equation:BinaryCE}) during training. Similarly, $\spadesuit$ denotes the exclusion of the second negative term, namely ``$(1-s_{ij})\log(1-\mathbf{p}_i^{\mathsf{T}}\mathbf{p}_j)$'', from Eq.~(\ref{Equation:BinaryCE}) when model training goes on.}
    \begin{tabular}{p{1.2cm}<{\centering}| p{0.8cm}<{\centering} p{0.8cm}<{\centering} p{0.8cm}<{\centering} p{0.8cm}<{\centering} p{0.8cm}<{\centering} |p{0.8cm}<{\centering} p{0.8cm}<{\centering} p{0.8cm}<{\centering} p{0.8cm}<{\centering}|p{0.8cm}<{\centering}}
    \toprule[1.5pt]
    \toprule
    M-(\textcolor{red}{\#}) &$\mathcal{L}^{ce}$&$\mathcal{L}^{adbc}$&$\mathcal{L}^{wdbc}$& $\mathcal{L}^{lab}$&$\mathcal{L}^{con}$& R$\rightarrow$C   & C$\rightarrow$S   & S$\rightarrow$P   & R$\rightarrow$S& Mean \\
    \midrule
    1     &\checkmark&    -     &    -     &    -     &    -     & 60.0  & 55.0  & 59.5  & 50.1  & 56.2  \\
    2     &\checkmark&\checkmark&    -     &    -     &    -     & 78.0  & 67.6  & 71.0  & 69.5  & 71.5  \\
    3     &\checkmark&    -     &\checkmark&    -     &    -     & 78.1  & 68.8  & 70.8  & 70.8  & 72.1  \\
    4     &\checkmark&\checkmark&\checkmark&    -     &    -     & 80.3  & 70.0  & 73.2  & 72.5  & 74.0  \\
    5     &\checkmark&    -     &    -     &\checkmark&    -     & 77.2  & 70.2  & 74.0  & 69.9  & 72.8  \\
    6    &\checkmark&    -     &    -     &    -     &\checkmark& 72.6  & 66.1  & 69.1  & 65.5  & 68.4  \\ % 11 -> 6
    7    &\checkmark&    -     &    -     &\checkmark&\checkmark& 78.3  & 71.5  & 74.3  & 71.0  & 73.8  \\ % 12 -> 7
    8     &\checkmark&\checkmark&    -     &\checkmark&    -     & 80.5  & 72.4  & 75.3  & 72.7  & 75.2  \\ % 6 -> 8
    9     &\checkmark&    -     &\checkmark&\checkmark&    -     & 81.0  & 73.1  & 76.0  & 73.1  & 75.8  \\ % 7 -> 9
    10     &\checkmark&\checkmark&\checkmark&\checkmark&    -     & 81.6  & 73.2  & 75.9  & 73.8  & 76.1  \\ % 8 -> 10
    11    &\checkmark&\checkmark&\checkmark&\checkmark&\checkmark& 82.2  & 73.4  & 76.3  & 74.3  & 76.5  \\ % 9 -> 11
    12    &\checkmark&$\dagger$ &$\dagger$ &\checkmark&\checkmark& 81.5  & 72.1  & 75.3  & 72.5  & 75.4  \\ % 10 -> 12
    13    &\checkmark&$\dagger$ &$\dagger$ &    -     &    -     & 78.6  & 68.5  & 72.9  & 70.4  & 72.6  \\
    14    &\checkmark&\checkmark &$\ddagger$ &\checkmark&\checkmark& 82.0  & 72.9  & 75.7  & 73.2  & 75.9  \\ 
    15    & \checkmark & $\clubsuit$ & $\clubsuit$ & \checkmark & \checkmark & 81.3  & 72.8  & 75.0  & 73.1  & 75.6  \\
    16    & \checkmark & $\spadesuit$ & $\spadesuit$ & \checkmark & \checkmark & 79.8  & 72.0  & 75.5  & 72.4  & 74.9  \\
    \bottomrule
    \bottomrule[1.5pt]
    \end{tabular}%
  \label{Table:Ablation}%
\end{table*}%

\subsection{Comparison with state-of-the-arts}

We compare the classification performance of our proposed G-ABC algorithm to that of previous state-of-the-art SSDA algorithms, including {\bf S+T}~\cite{saito2019semi}, {\bf DANN}~\cite{saito2019semi}, {\bf MME}~\cite{saito2019semi}, {\bf UODA}~\cite{qin2021contradictory}, {\bf Meta-MME}~\cite{li2020online}, {\bf BiAT}~\cite{2020Bidirectional}, {\bf APE}~\cite{kim2020attract}, {\bf ELP}~\cite{ELP}, {\bf PAC}~\cite{mishra2021surprisingly}, {\bf Relaxed-cGAN}~\cite{luo2021relaxed}, {\bf DECOTA}~\cite{YangLuyu2020DCwT}, \textbf{ECACL-T}~\cite{li2021ecacl},  \textbf{ECACL-P}~\cite{li2021ecacl}, ${\text{\textbf{S}}}^{\text{\textbf{3}}}{\text{\textbf{D}}}$ \cite{R3C2}, \textbf{UODAv2} \cite{R3C1}, \textbf{MCL}~\cite{R3C3}, \textbf{CLDA}~\cite{singh2021clda} and {\bf CDAC}~\cite{li2021cross}. Note that \textbf{S+T} refers to an approach that trains the adaptation model solely with supervision on labeled samples from both domains, whilst \textbf{DANN} refers to the method presented in~\cite{ganin2016domain}, but additionally applies a standard cross-entropy loss on a few labeled samples in the target domain.

\noindent\textbf{{On DomainNet.}}
In order to highlight the advantages of the proposed algorithm, we compare our G-ABC strategy to numerous existing alternatives on the DomainNet benchmark. Table~\textcolor{black}{\ref{Table:DomainNet}} presents the results of this dataset benchmark utilizing 1-shot and 3-shot settings with AlexNet and ResNet-34 as the corresponding backbone networks. As demonstrated, our proposed G-ABC method achieves more average performance gains than all existing approaches in the majority of DomainNet adaptation cases. Specifically, G-ABC improves the prior best-performing SSDA algorithm, i.e., {\bf ECACL-T} and {\bf DECOTA}, by mean accuracy margins of 2.3\% and 3.1\% while employing AlexNet and ResNet-34 as the backbones, respectively, for all adaptation scenarios under the 1-shot setting. In addition, the proposed method outperforms competing approaches in most of the adaptation scenarios defined on DomainNet with a 3-shot setting by outperforming the best available results (accuracy of 57.7\%  and 76.5\% in {\bf ECACL-T} and \textbf{MCL}) by 1.88\% and 2.08\% on average, when AlexNet and ResNet-34 serve as the backbone networks, respectively. These results demonstrate the effectiveness of our algorithm in dealing with SSDA tasks on DomainNet.
 
\noindent\textbf{{On Office-Home.}}
To validate the feasibility of the proposed G-ABC algorithm in SSDA, we also compare the results of our method to those of earlier methods on Office-Home. Similar to prior baselines~\cite{saito2019semi, qin2021contradictory, kim2020attract, YangLuyu2020DCwT, li2021cross}, we conduct experiments on this dataset under the 3-shot setting, AlexNet and ResNet-34 as the backbones, and all 12 adaptation scenarios for Office-Home. Table~\textcolor{black}{\ref{Table:OfficeHome}} illustrates the classification accuracy of each adaptation scenario and the average performance of the proposed G-ABC algorithm on Office-Home, respectively. As demonstrated, the proposed method achieves the best average classification performance when utilizing the ResNet-34 backbone, and the accuracy surpasses the best baseline \textbf{DECOTA} by significant margins of 1.49\%. 

\noindent\textbf{{On Office-31.}}
Aiming at further confirming the efficacy of the proposed G-ABC method, we conduct experiments comparing to the existing state-of-the-art approaches using the smallest benchmark dataset, namely Office-31. In order to retain consistency with prior approaches and to assure a fair comparison, we only use AlexNet as the backbone of this work. Table~\textcolor{black}{\ref{Table:Office31}} shows that our method obtains the highest classification performance in both ``W $\rightarrow$ A'' and ``D $\rightarrow$ A'' cases, with an average accuracy of 66.81\% (+3.21\%) under the 1-shot setting and 72.02\% (+0.52\%) under the 3-shot setting. In other words, our strategy outperforms the existing best-performing baseline, \textbf{CLDA}, by significant average accuracy margins in all adaptation scenarios, indicating the improved effectiveness of the proposed method on this dataset.

\noindent
\textbf{\textit{Discussion}.} It appears that superior performance gains have been observed on the DomainNet dataset compared to Office-Home. This is because Office-Home is a relatively simpler SSDA benchmark dataset. As shown in Table~\textcolor{black}{\ref{Table:OfficeHome}}, it is evident that prior state-of-the-art (SOTA) methods have reached their performance limits on this dataset. Similar to these approaches, this saturation in performance makes it challenging for our method to achieve significant improvements compared to existing SOTA methods.
In contrast, DomainNet has a larger domain shift, which poses challenges for domain adaptation methods and offers more room for improvement. As illustrated in Table~\textcolor{black}{\ref{Table:DomainNet}}, our method demonstrates more notable gains on DomainNet, as it is designed to effectively handle such complex domain shifts.

\subsection{Ablation Study}
\begin{figure}[t]
    \centering
    \includegraphics[width=0.75\linewidth]{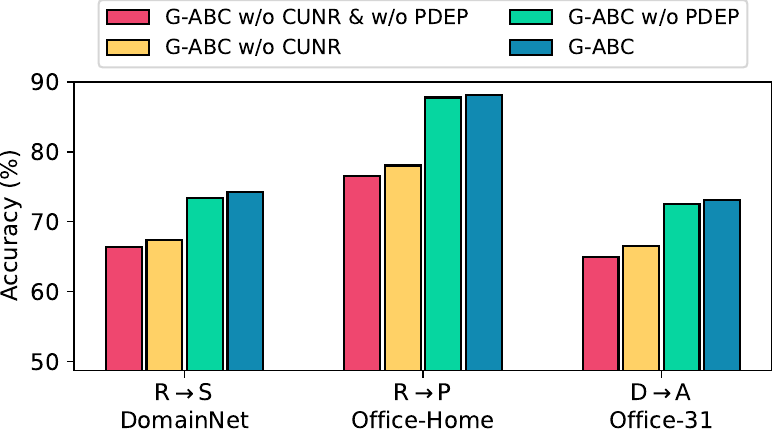}
    \caption{The impact of removing CUNR and PDEP on performance during graph construction. The experiments are performed on adaptation scenarios of ``R$\rightarrow$S'' on DomainNet, ``R$\rightarrow$P'' on Office-Home, and ``D$\rightarrow$A'' on Office-31, respectively. All of them are conducted under a 3-shot setup, with the first two using ResNet-34 as network backbones, while the latter uses AlexNet.}
    \label{Figure:Removal-CUNR-and-PDEP}
\end{figure}

We conduct extensive experiments to individually confirm the efficacy of each component of our proposed G-ABC approach. Specifically, Table~\textcolor{black}{\ref{Table:Ablation}} shows the main ablation study results, where all experiments are performed in four adaptation scenarios on DomainNet using ResNet-34 as the backbone under the 3-shot setup. Furthermore, Fig.~\ref{Figure:Confusion} illustrates the evaluation of Adaptive Betweenness Clustering in both ADBC and WDBC, while Fig~\ref{Figure:Removal-CUNR-and-PDEP} demonstrates the impact of removing CUNR and PDEP on graph construction and their influence on the model performance.
 
\noindent
{\bf Effectiveness of ADBC and WDBC.} To determine the effectiveness of $\mathcal{L}^{adbc}$ and $\mathcal{L}^{wdbc}$ proposed in our method, we first train the model using only labeled samples from both domains, serving as the baseline being depicted in row M-(\textcolor{red}{1}) of Table~\textcolor{black}{\ref{Table:Ablation}}. According to Table~\textcolor{black}{\ref{Table:Ablation}}, training the model with both ADBC and WDBC delivers greater classification performance gains than training the model with simply one of both. It can be observed that row M-(\textcolor{red}{4}) in Table~\textcolor{black}{\ref{Table:Ablation}} increases the baseline by an average of 17.9\%, while the accuracy rates in row M-(\textcolor{red}{2}) and row M-(\textcolor{red}{3}) can only exceed the baseline by 15.4\% and 16.0\%, respectively, thereby confirming the validity of the ADBC and WDBC stages. In addition, when row M-(\textcolor{red}{5}) of Table~\textcolor{black}{\ref{Table:Ablation}} is considered as another baseline, a similar situation can be observed when contrasting among row M-(\textcolor{red}{8}), row M-(\textcolor{red}{9}) and row M-(\textcolor{red}{10}).

\noindent
{\bf Effectiveness of Adaptive Betweenness Clustering.} In order to further understand the role of the proposed clustering method, Adaptive Betweenness Clustering (ABC), in ADBC and WDBC, we employed the Class-wise Similarity Score (CSS), which is illustrated in detail in Sec. \ref{SubSection:Implementation}. The CSS is used to measure the similarity in predictions between unlabeled target samples and labeled source or target domains. As depicted in the heatmaps of Fig.~\textcolor{black}{\ref{Figure:Confusion}}, the CSS scores between the same classes progressively increase, while those between different classes decrease throughout the model training process. This observation indicates that unlabeled target samples have a tendency to cluster together with labeled source and target data from the same class in the feature space, while samples from different classes become more distant from each other. This demonstrates the efficacy of Adaptive Betweenness Clustering in facilitating semantic propagation, thereby raising the performance of ADBC and WDBC. To further validate the effectiveness of the Adaptive Betweenness Clustering loss, we also performed validation experiments on the two terms of the loss presented in Eq.~(\ref{Equation:BinaryCE}), namely the first positive term ``$s_{ij}\log(\mathbf{p}_i^{\mathsf{T}}\mathbf{p}_j)$'' and the second negative term ``$(1-s_{ij})\log(1-\mathbf{p}_i^{\mathsf{T}}\mathbf{p}_j)$''. According to Table~\textcolor{black}{\ref{Table:Ablation}}, it can be observed that row M-(\textcolor{red}{15}), M-(\textcolor{red}{16}), and M-(\textcolor{red}{7}) correspond to the removal of individual terms or both terms from Eq.~(\ref{Equation:BinaryCE}). By comparing these results with the classification performance of the full model indicated by row M-(\textcolor{red}{11}), it can be seen that removing either one or both terms leads to a decrease in model performance, with a more pronounced effect when both terms are removed. This finding indirectly verifies the effectiveness of the adaptive betweenness clustering loss.
 
\noindent
{\bf Effectiveness of CUNR and PDEP.} To confirm the impact of CUNR and PDEP during graph construction, we present the results of removing each of them individually, as well as both, on three SSDA benchmarks. As shown in Fig.~\textcolor{black}{\ref{Figure:Removal-CUNR-and-PDEP}}, the comparison between the full model ``G-ABC'' and its variants ``G-ABC w/o CUNR'' or ``G-ABC w/o PDEP'' demonstrates that removing either CUNR or PDEP leads to a significant decline in the model's overall performance. Moreover, when both are eliminated, referred to as ``G-ABC w/o CUNR \& w/o PDEP'', the model's performance reaches its lowest point. This demonstrates that CUNR or PDEP effectively eliminates noisy connectivity while constructing a reliable graph structure, thereby enhancing the model's performance. In particular, CUNR provides a greater performance improvement than PDEP. This is because unlabeled target samples with lower confidence, which would be removed by CUNR during training, not only negatively impact samples of the same class but also affect samples from different classes, resulting in more substantial harm to the model's performance.

\noindent
{\bf Effectiveness of Self-training.} Examining the necessity of $\mathcal{L}^{lab}$, we should use experiments in row M-(\textcolor{red}{4})  of Table~\textcolor{black}{\ref{Table:Ablation}} in which the model is trained with $\mathcal{L}^{ce}$, $\mathcal{L}^{adbc}$ and $\mathcal{L}^{wdbc}$ as the baseline. As shown in Table~\textcolor{black}{\ref{Table:Ablation}}, the average performance of row M-(\textcolor{red}{10}) with additional $\mathcal{L}^{lab}$ loss is 2.1\% more than the baseline, indicating the necessity of this component for our proposed G-ABC approach.
 
\noindent
{\bf Effectiveness of Consistency Training.}
Table~\textcolor{black}{\ref{Table:Ablation}} indicates the effectiveness of consistency training as well. 
Comparing row M-(\textcolor{red}{10}) (or row M-(\textcolor{red}{6})) with row M-(\textcolor{red}{11}) (or row M-(\textcolor{red}{7})), it is evident that consistency training for all unlabeled target data is beneficial. 

\noindent
{\bf Effectiveness of Pseudo-label Selection.}
In order to explore the impact of pseudo-label selection, we omit the pseudo-labeled target samples applying to Eq.~(\textcolor{black}{\ref{Equation:WdbcLoss}}). Table~\textcolor{black}{\ref{Table:Ablation}} demonstrates that in the comparison between rows M-(\textcolor{red}{11}) and M-(\textcolor{red}{14}), the average accuracy of row M-(\textcolor{red}{14}) is 0.6\% less than that of row M-(\textcolor{red}{11}), revealing that pseudo-label selection is also effective for enhancing the performance of the model.
 
\noindent{\bf Effectiveness of Sample Perturbation.} We propose to introduce perturbations into unlabeled target samples on both ADBC and WDBC. According to Table~\textcolor{black}{\ref{Table:Ablation}}, by comparing row M-(\textcolor{red}{4}) and row M-(\textcolor{red}{11}) with row M-(\textcolor{red}{13}) and row M-(\textcolor{red}{12}), respectively, we observe that the performance of the mean accuracy decreases by 1.4\% and 1.1\%, respectively, indicating that it is necessary to include perturbations in our model training. 

\subsection{Further Analysis}

\begin{figure*}[htbp]
    \centering
    \begin{minipage}{0.24\textwidth}
    \includegraphics[width=\linewidth]{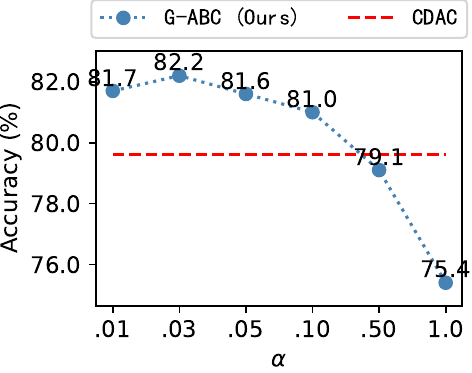}
    \caption*{\quad\quad(a)}
    \end{minipage}
    \hfill
    \begin{minipage}{0.24\textwidth}
    \includegraphics[width=\linewidth]{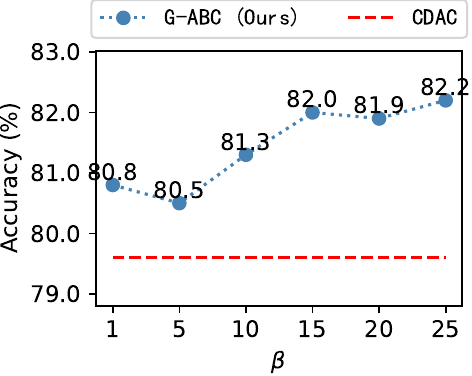}
    \caption*{\quad\quad(b)}
    \end{minipage}
    \hfill
    \begin{minipage}{0.24\textwidth}
    \includegraphics[width=\linewidth]{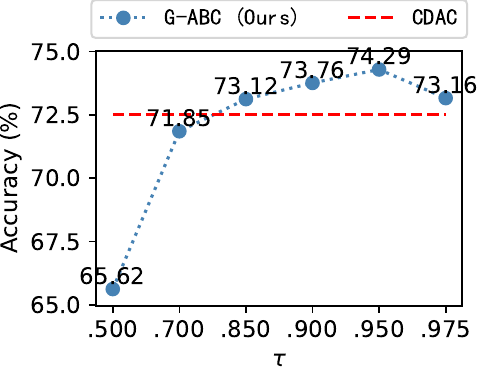}
    \caption*{\quad\quad(c)}
    \end{minipage}
    \hfill
    \begin{minipage}{0.24\textwidth}
    \includegraphics[width=\linewidth]{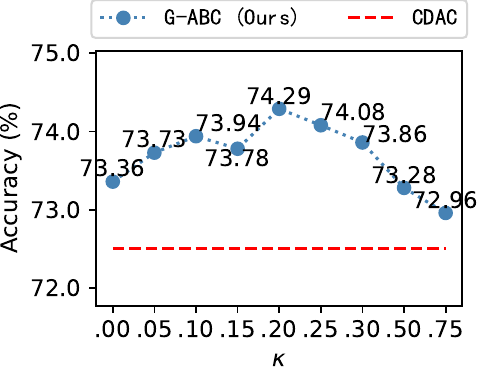}
    \caption*{\quad\quad(d)}
    \end{minipage}
    \caption{Sensitivity with respect to hyper-parameters $\alpha$, $\beta$, $\tau$ and $\kappa$. The experiments are performed on DomainNet under the setting of the 3-shot using the ResNet-34 backbone, where (a) and (b) are with ``R$\rightarrow$C'', and (c) and (d) are with ``R$\rightarrow$S''. ``\textbf{CDAC}''~\cite{li2021cross} is the best-performing baseline method in both adaptation scenarios. As illustrated, our method is not highly sensitive to changes in the hyper-parameters $\alpha$, $\beta$, $\tau$ and $\kappa$. This is because, over a wide range, our G-ABC approach outperforms the baseline method ``\textbf{CDAC}'' significantly for all four hyper-parameters.}
    \label{Figure:Ablation}%
\end{figure*}

We also investigate the hyper-parameter sensitivity to the loss weights $\alpha$ and $\beta$, the similarity threshold $\kappa$, as well as the hyper-parameter reasonability with respect to the confidence threshold $\tau^{\prime}$.
In addition, we visualize the feature distributions across domains using t-SNE~\cite{maaten2008visualizing}.
 
\noindent{\bf Hyper-parameter Sensitivity to $\alpha$ and $\beta$.} In Fig.~\textcolor{black}{\ref{Figure:Ablation}}(a) and (b), we highlight the influence of $\alpha$ and $\beta$. It can be observed that when $\alpha=0.03$ and $\beta=25.0$, the trained model achieves the highest performance in image classification. However, the accuracy decreases significantly when they are adjusted further from the optimal value. 
By introducing $g_i^j$ in Eqs. (\textcolor{black}{\ref{Equation:WdbcLoss}}) and (\textcolor{black}{\ref{Equation:AdbcLoss}}) to remove graph nodes with noise sample connectivity, a larger proportion of unlabeled target samples do not actually participate in model updates but do contribute to the calculation of $\mathcal{L}_{abc}$, resulting in a smaller scale of $\mathcal{L}_{abc}$. Therefore, setting $\beta$ to a higher value, i.e., 25.0, is advantageous for balancing the influence of $\mathcal{L}_{abc}$ and other loss items during model updates.
 
\noindent
{\bf Hyper-parameter Sensitivity to $\tau$ and $\kappa$ in CUNR and PDEP.} We also conduct experiments to assess the sensitivity of our method to the hyperparameters $\tau$ and $\kappa$. Fig.~\textcolor{black}{\ref{Figure:Ablation}}(c) and (d) illustrate that the classification performance of the model achieves best when $\tau$ and $\kappa$ are set to 0.95 and 0.20, respectively; however, changing either of these parameters, especially on $\tau$ to less than 0.90 and $\kappa$ to greater than 0.50, results in a decline in accuracy. This is due to the fact that lighter CUNR causes a greater number of non-confident unlabeled target samples to be preserved in the graph, whereas higher PDEP causes an excessive number of node removals inside the graph, resulting in unreliable knowledge transfer between target samples.

\begin{figure}[t!]
    \centering
    \begin{minipage}{0.24\textwidth}
    \includegraphics[width=\linewidth]{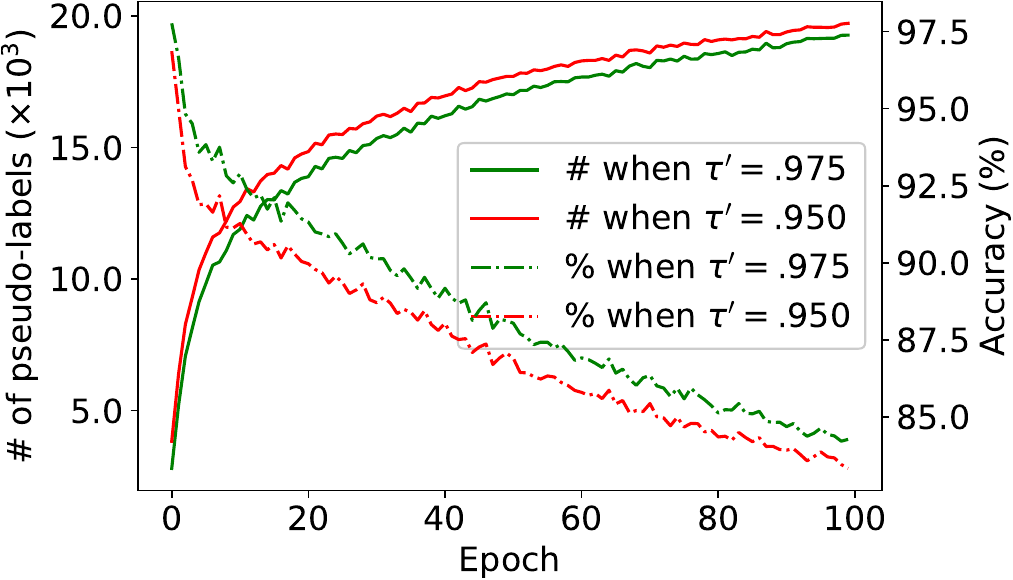}
    \caption*{(a) ``R$\rightarrow$S'' on DomainNet}
    \end{minipage}
    \hfill
    \begin{minipage}{0.24\textwidth}
    \includegraphics[width=\linewidth]{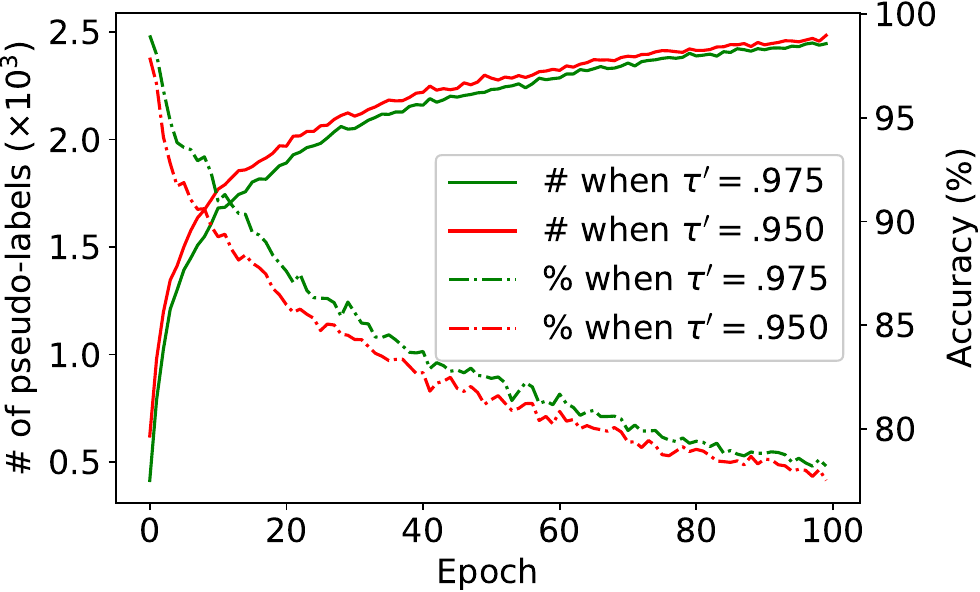}
    \caption*{(b) ``D$\rightarrow$A'' on Office-31}
    \end{minipage}
    \caption{The evolution of the numbers ($\#$) and the accuracy  ($\%$) of the pseudo-labels over epochs while varying the confidence threshold ${\tau}^{\prime}$. The experiments are performed with (a): ``R$\rightarrow$S'' on DomainNet using ResNet-34, and (b): ``D$\rightarrow$A'' on Office-31 using AlexNet, both under the 3-shot setup.}
    \label{Figure:PseudoLabel}
\end{figure}

\noindent
{\bf Hyper-parameter Rationality to $\tau^{\prime}$.}  We conduct additional experiments to prove the validity of setting ${\tau}^{\prime}$ to 0.975 as opposed to 0.95 by plotting variations in the quantity and accuracy of pseudo-labels for target samples whose expected probability are greater than ${\tau}^{\prime}$. As depicted in Fig.~\textcolor{black}{\ref{Figure:PseudoLabel}}, when ${\tau}^{\prime}$ is set to 0.95, $\mathcal{D}_{pu}$ can collect significantly more pseudo-labeled target samples from $\mathcal{D}_{u}$. However, the large amount of noise contained in the pseudo-labels will also bring challenges to the model. This proves setting $\tau^{\prime}$ to 0.975 is better than 0.95.

\begin{figure}[h]
    \centering
    \begin{minipage}{0.24\textwidth}
        \includegraphics[width=\linewidth]{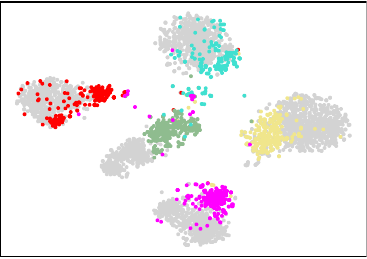}
        \caption*{(a) ``R $\rightarrow$ S'' on DomainNet}
    \end{minipage}
    \hfill
    \begin{minipage}{0.24\textwidth}
        \includegraphics[width=\linewidth]{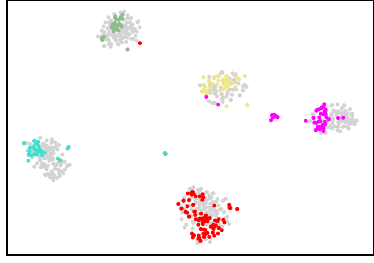}
        \caption*{(b) ``R $\rightarrow$ P'' on Office-Home}
    \end{minipage}
    \caption{Feature visualization using t-SNE. We randomly choose five classes with (a): ``R $\rightarrow$ S'' on DomainNet, and (b): ``R $\rightarrow$ P'' on Office-Home, respectively,  with the ResNet-34 backbone and the 3-shot setup. Herein, data points in grey represent source samples, while brightly colored examples are from the target domain. The red, lightblue, purple, green, and yellow represent the categories of ``Axe'', ``Bird'', ``Fence'', ``Shoe'', and ``Truck'' on DomainNet, while these colors correspond to the categories of ``Batteries'', ``Calendar'', ``Flowers'', ``Glasses'', and ``Monitor'' within the  Office-Home dataset.}
    \label{Figure:TSNE}
\end{figure}

\noindent
{\bf Feature Visualization.} Using t-SNE for feature visualization, we present the feature distributions obtained by the proposed method for both domains in Fig.~\textcolor{black}{\ref{Figure:TSNE}}. It can be observed that in the feature space, the learned features from different domains that belong to the same class are mapped nearby and clustered together, whereas those from distinct categories are significantly separated. This implies that the model trained using the proposed G-ABC approach is capable of producing domain-invariant and discriminative target features, thus contributing to the improved performance of the SSDA task.
 
\section{Conclusions}
\label{Section:Conclusions}

This paper presents a novel SSDA method named Graph-based Adaptive Betweenness Clustering for achieving categorical domain alignment. It facilitates cross-domain semantic alignment by enforcing semantic transfer from labeled source and target data to unlabeled target samples. In this approach, a heterogeneous graph is first constructed to represent pairwise relationships between labeled examples from both domains and unlabeled target samples. Then, two strategies including Confidence Uncertainty based Node Removal and Prediction Dissimilarity based Edge Pruning are proposed to refine the connectivity in the graph to alleviate the influence of noisy edges. Provided with the refined graph, we present adaptive betweenness clustering to accomplish semantic transfer across domains with semantic propagation from labeled source or target examples to unlabeled samples on the target domain. Extensive experimental results as well as comprehensive analysis performed well on three benchmark datasets demonstrate the superiority of our proposed method, achieving new state-of-the-art results.

\bibliography{IEEEabrv,mybib}{}
\bibliographystyle{IEEEtran}

\begin{IEEEbiography}[{\includegraphics[width=1in,height=1.25in,clip,keepaspectratio]{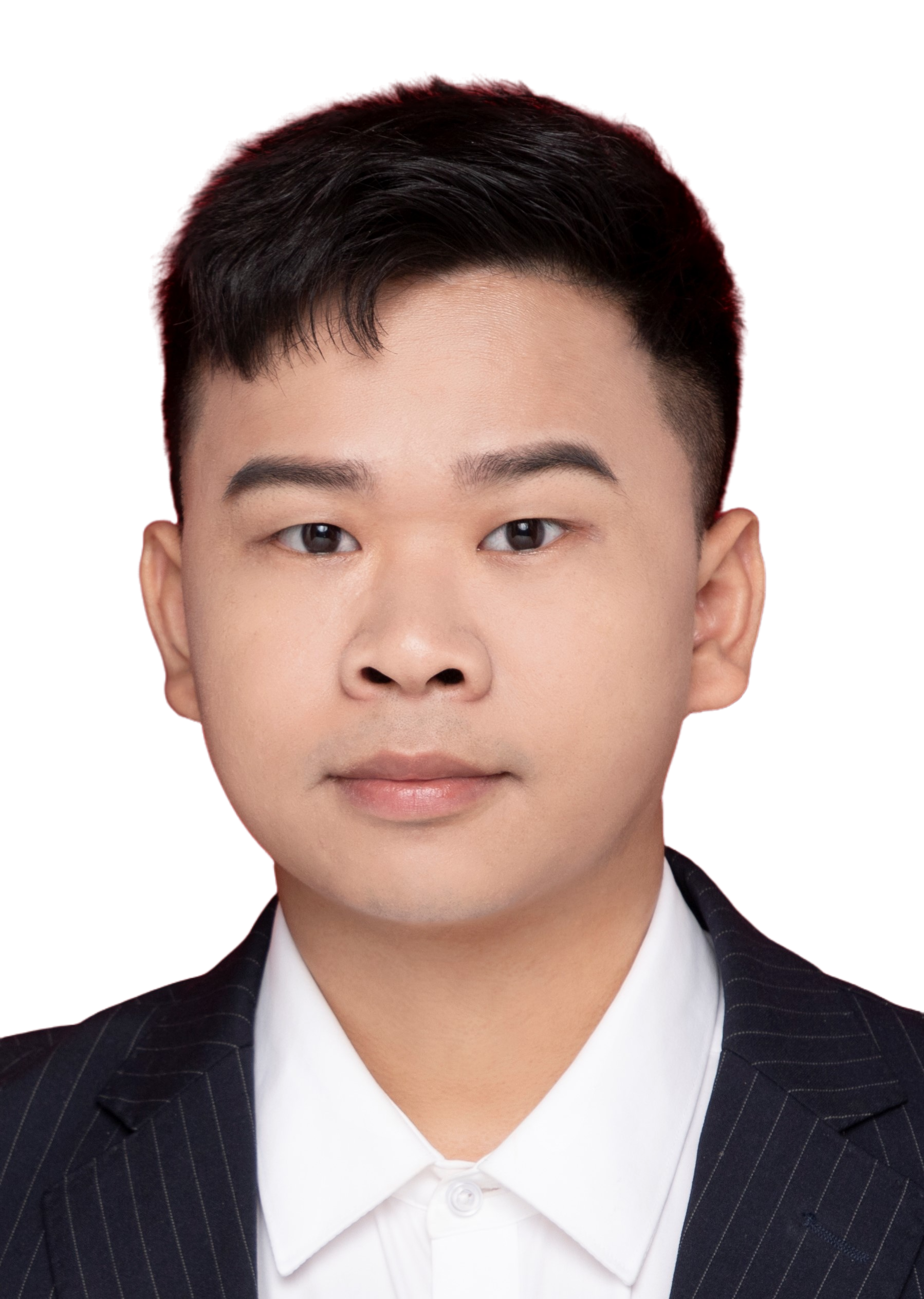}}]{Jichang Li} is currently pursuing the PhD degree in the Department of Computer Science, The University of Hong Kong. In 2020, he received the M.Eng. degree from the School of Computer Science and Technology, South China University of Technology. His current research interests include computer vision and deep learning.
\end{IEEEbiography}

\begin{IEEEbiography}[{\includegraphics[width=1in,height=1.25in,clip,keepaspectratio]{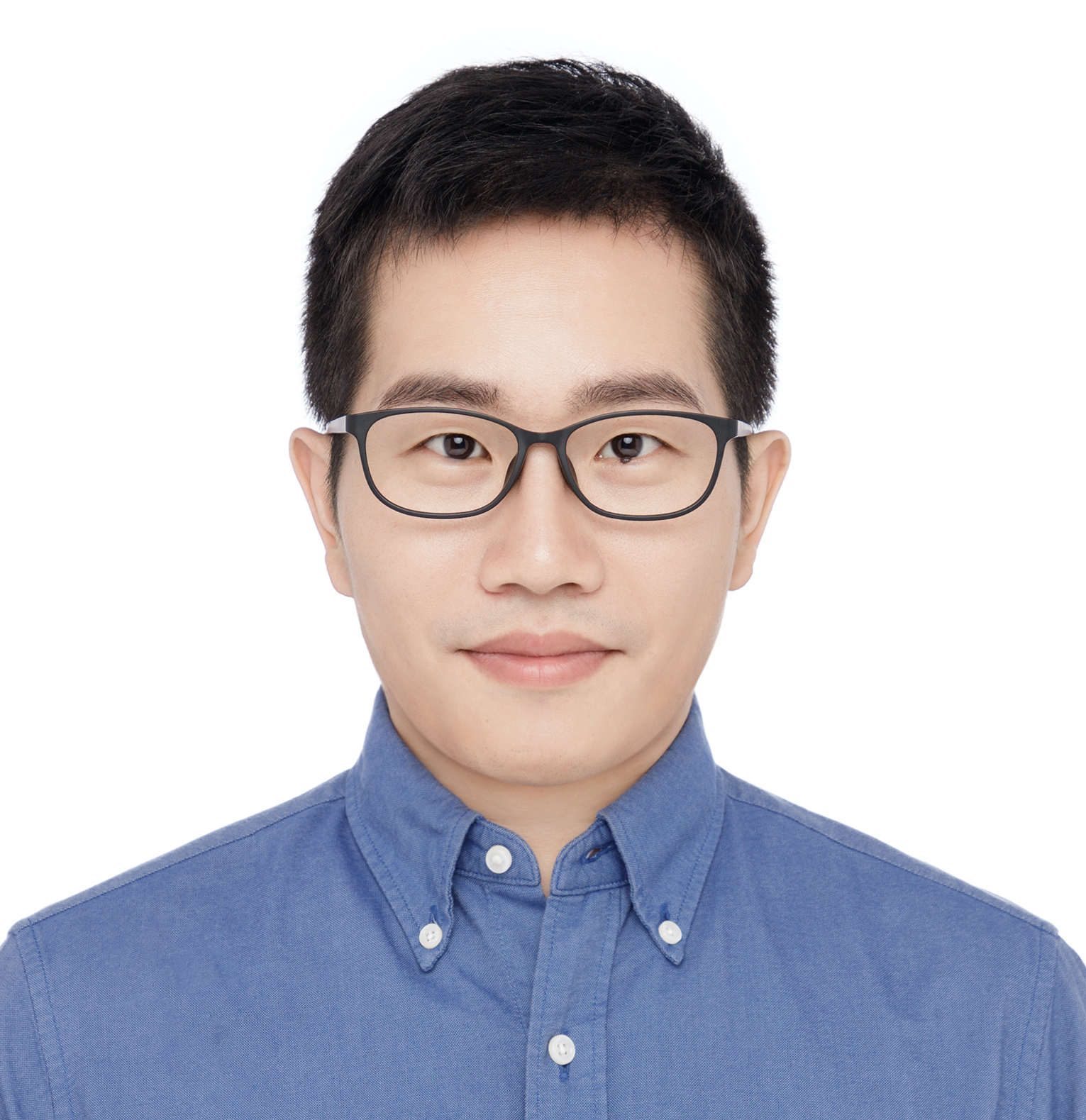}}]{Guanbin Li} (M'15) is currently an associate professor in the School of Computer Science and Engineering, Sun Yat-sen University. He received his PhD degree from The University of Hong Kong in 2016. His current research interests include computer vision, image processing, and deep learning. He is a recipient of ICCV 2019 Best Paper Nomination Award. He has authorized and co-authorized on more than 100 papers in top-tier academic journals and conferences. He serves as an area chair for the conference of VISAPP. He has been serving as a reviewer for numerous academic journals and conferences such as TPAMI, IJCV, TIP, TMM, TCyb, CVPR, ICCV, ECCV and NeurIPS.
\end{IEEEbiography}
 
\begin{IEEEbiography}[{\includegraphics[width=1in,height=1.25in,clip,keepaspectratio]{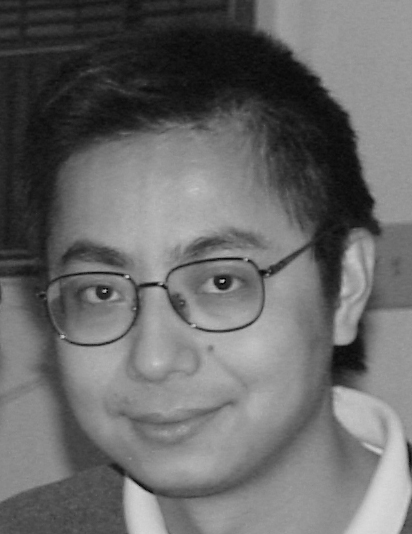}}]{Yizhou Yu} (M'10, SM'12, F'19) received the PhD degree from the University of California at Berkeley in 2000. He is a chair professor with the University of Hong Kong, and was a faculty member with the University of Illinois at Urbana-Champaign for twelve years. He is a recipient of 2002 US National Science Foundation CAREER Award and ACCV 2018 Best Application Paper Award. He has served on the editorial board of IEEE Transactions on Pattern Analysis and Machine Intelligence, IEEE Transactions on Image Processing, and IEEE Transactions on Visualization and Computer Graphics. He has also served on the program committee of many leading international conferences, including CVPR, ICCV, and SIGGRAPH. His current research interests include AI foundation models, AI based multimedia content generation, AI for medicine, and computer vision.
\end{IEEEbiography} 

\end{document}